\pdfoutput=1

\documentclass[11pt]{article}

\usepackage[final]{acl} 

\usepackage{times}
\usepackage{latexsym}

\usepackage{amsmath}
\usepackage{amssymb}
\usepackage{mathtools}
\usepackage{amsthm}

\usepackage{makecell}
\usepackage{multirow}
\usepackage{subfig}
\usepackage{subfiles}
\usepackage{xcolor,colortbl}
\usepackage{centernot}
\usepackage{booktabs}
\usepackage{pifont}

\usepackage{enumitem}
\usepackage[most]{tcolorbox}

\usepackage[textsize=tiny]{todonotes}
\newtcbox{\hlprimarytab}{on line, rounded corners, box align=base, colback=red!10,colframe=white,size=fbox,arc=3pt, before upper=\strut, top=-2pt, bottom=-4pt, left=-2pt, right=-2pt, boxrule=0pt}
\newtcbox{\hlsecondarytab}{on line, box align=base, colback=blue!10,colframe=white,size=fbox,arc=3pt, before upper=\strut, top=-2pt, bottom=-4pt, left=-2pt, right=-2pt, boxrule=0pt}
\newtcbox{\hlthirdtab}{on line, box align=base, colback=gray!10,colframe=white,size=fbox,arc=3pt, before upper=\strut, top=-2pt, bottom=-4pt, left=-2pt, right=-2pt, boxrule=0pt}
\newtcbox{\hlfourthtab}{on line, box align=base, colback=blue!20,colframe=white,size=fbox,arc=3pt, before upper=\strut, top=-2pt, bottom=-4pt, left=-2pt, right=-2pt, boxrule=0pt}

\newtcbox{\rtabone}{on line, rounded corners, box align=base, colback=red!10,colframe=white,size=fbox,arc=3pt, before upper=\strut, top=-2pt, bottom=-4pt, left=-2pt, right=-2pt, boxrule=0pt}
\newtcbox{\rtabtwo}{on line, rounded corners, box align=base, colback=red!20,colframe=white,size=fbox,arc=3pt, before upper=\strut, top=-2pt, bottom=-4pt, left=-2pt, right=-2pt, boxrule=0pt}
\newtcbox{\rtabthree}{on line, rounded corners, box align=base, colback=red!30,colframe=white,size=fbox,arc=3pt, before upper=\strut, top=-2pt, bottom=-4pt, left=-2pt, right=-2pt, boxrule=0pt}
\newtcbox{\rtabultra}{on line, rounded corners, box align=base, colback=red!60,colframe=white,size=fbox,arc=3pt, before upper=\strut, top=-2pt, bottom=-4pt, left=-2pt, right=-2pt, boxrule=0pt}

\newtcbox{\btabone}{on line, rounded corners, box align=base, colback=blue!10,colframe=white,size=fbox,arc=3pt, before upper=\strut, top=-2pt, bottom=-4pt, left=-2pt, right=-2pt, boxrule=0pt}
\newtcbox{\btabtwo}{on line, rounded corners, box align=base, colback=blue!20,colframe=white,size=fbox,arc=3pt, before upper=\strut, top=-2pt, bottom=-4pt, left=-2pt, right=-2pt, boxrule=0pt}
\newtcbox{\btabthree}{on line, rounded corners, box align=base, colback=blue!30,colframe=white,size=fbox,arc=3pt, before upper=\strut, top=-2pt, bottom=-4pt, left=-2pt, right=-2pt, boxrule=0pt}

\newcommand{\dashifted}{\raisebox{0.5\depth}{\tiny$\downarrow$}}

\newcommand{\uashifted}{\raisebox{0.5\depth}{\tiny$\uparrow$}}

\newcommand\mtiny[1]{\mbox{\tiny\ensuremath{#1}}}
\newcommand{\redone}[1]{{\mtiny{\rtabone{\dashifted{#1}}}}}
\newcommand{\redtwo}[1]{{\mtiny{\rtabtwo{\dashifted{#1}}}}}
\newcommand{\redthree}[1]{{\mtiny{\rtabthree{\dashifted{#1}}}}}
\newcommand{\redultra}[1]{{\mtiny{\rtabultra{\dashifted{#1}}}}}
\newcommand{\blueone}[1]{{\mtiny{\btabone{\uashifted{#1}}}}}
\newcommand{\bluetwo}[1]{{\mtiny{\btabtwo{\uashifted{#1}}}}}
\newcommand{\bluethree}[1]{{\mtiny{\btabthree{\uashifted{#1}}}}}

\usepackage[T1]{fontenc}

\usepackage[utf8]{inputenc}

\usepackage{microtype}

\usepackage{inconsolata}

\providecommand{\todo}[1]{
    
}

\providecommand{\danqi}[1]{

}

\newcommand\ab[1]{}
\newcommand{\adithya}[1]{}
\newcommand{\dan}[1]{}

\newcommand\ti[1]{\textit{#1}}

\newcommand\tf[1]{\textbf{#1}}

\renewcommand{\paragraph}[1]{\vspace{0.2cm}\noindent\textbf{#1}}

\title{The Heuristic Core: Understanding Subnetwork Generalization \\ in Pretrained Language Models}

\author{Adithya Bhaskar\hspace{-4em} \And Dan Friedman \\Princeton Language and Intelligence (PLI), Princeton University\\
\texttt{adithyab@princeton.edu}\hspace{1.5em} \texttt{\{dfriedman, danqic\}@cs.princeton.edu}\And \hspace{-4em}Danqi Chen}

\begin{document}
\maketitle
\begin{abstract}
Prior work has found that pretrained language models (LMs) fine-tuned with different random seeds can achieve similar in-domain performance but generalize differently on tests of syntactic generalization. 
In this work, we show that, even within a single model, we can find multiple subnetworks that perform similarly in-domain, but generalize vastly differently.
To better understand these phenomena, we investigate if they can be understood in terms of ``competing subnetworks'': the model initially represents a variety of distinct algorithms, corresponding to different subnetworks, 
and generalization occurs when it ultimately converges to one.
This explanation has been used to account for generalization in simple algorithmic tasks (``grokking'').
Instead of finding competing subnetworks, we find that all subnetworks---whether they generalize or not---share a set of attention heads, which we refer to as the \ti{heuristic core}.
Further analysis suggests that these attention heads emerge early in training and compute shallow, non-generalizing features.
The model learns to generalize by incorporating additional attention heads, which depend on the outputs of the ``heuristic'' heads to compute higher-level features.
Overall, our results offer a more detailed picture of the mechanisms for syntactic generalization in pretrained LMs.\footnote{We release our code publicly at \url{https://github.com/princeton-nlp/Heuristic-Core}.}

\end{abstract}
\section{Introduction}
\label{sec:intro}
\begin{figure}[t]
    \centering
    \includegraphics[width=\linewidth]{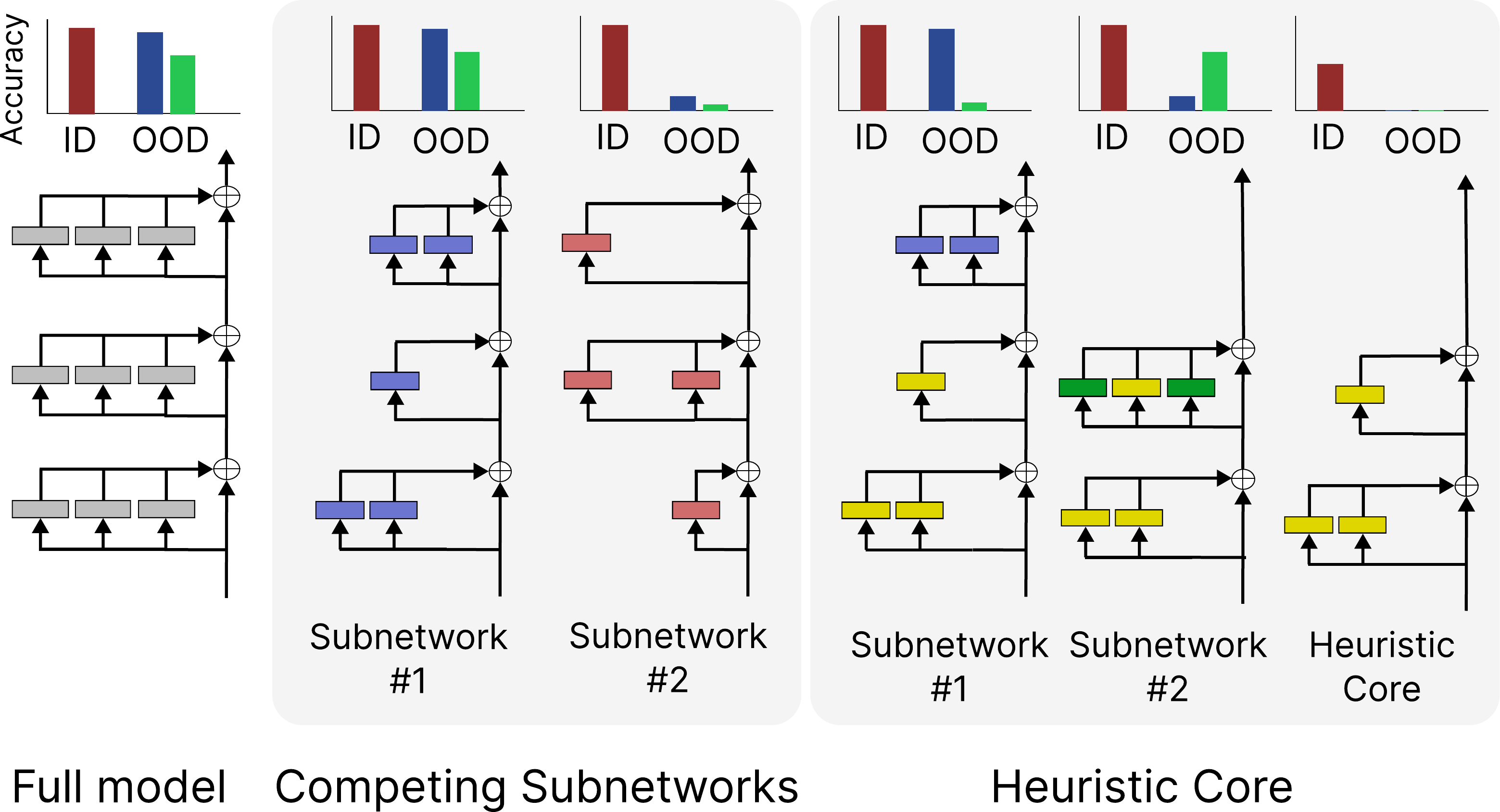}
    \label{fig:teaser}
    \caption{
    We find different subnetworks in a pretrained LM that achieve similar in-domain performance but generalize differently.
    Prior work has explained similar generalization phenomena in synthetic tasks in terms of distinct subnetworks that compete during training.
    We instead find evidence of a \ti{heuristic core}: a set of attention heads that appear in all generalizing subnetworks but, on their own, do not generalize.
    }
\end{figure} 
A central question in machine learning is to understand how models generalize from data that supports different possible solutions.
~\citet{bertsofafeather} investigated this question in the context of pretrained language models (LMs) like BERT~\citep{bert} trained on natural language inference~\citep[NLI; ][]{williams2018broad}, showing that models trained with different random seeds perform very similarly on in-domain (ID) evaluation sets but generalize very differently to adversarial out-of-domain (OOD) evaluation sets.
However, we have a limited understanding of how any individual model learns to generalize.

In this work, we use subnetwork analysis to better understand how pretrained LMs generalize on NLP tasks.
Specifically, we use structured pruning~\cite{structuredpruning,cofi} to isolate various subnetworks---subsets of attention heads and MLP layers---that approximate the behavior of the full model.
Our main finding is, that even within a single model, there exist multiple subnetworks that all match the model's performance closely on in-domain evaluation sets, but exhibit vastly different generalization.
This result has meaningful consequences for practitioners---for example, it underscores the importance of OOD evaluation of pruning methods. It also raises several questions about the mechanisms underlying generalization, which we proceed to investigate.

One possible explanation for our results is that the model initially consists of a variety of disjoint subnetworks, corresponding to distinct possible solutions for the task; the model ultimately generalizes if it converges to a subnetwork that generalizes.
This model of generalization has been used to explain the ``grokking'' phenomenon on toy algorithmic tasks, whereby a neural network initially overfits a training set---but, after continued training, suddenly undergoes a phase shift to perfect generalization~\citep{hiddenprogress,progressmeasures,grokkingexplaindeepmind}.
We refer to this as the \ti{Competing Subnetworks} explanation, following~\citet{taleoftwocircuits}.
Our setting bears a similarity to grokking, originally documented by~\citet{smallexamplesgen}: models converge early on in-domain evaluations and only start to generalize after training for additional epochs.
Therefore, we test if generalization in this case also arises from competition between disjoint subnetworks.

\begin{table*}[t]
    \centering
    \small
    \begin{tabular}{lccc}
    \toprule
    \textbf{Subcase} & \textbf{Name} & \textbf{Abbreviation} & \textbf{Example}\\
    \midrule
    \textbf{\textit{Lexical Overlap (LO)}} & & & \\
    \quad Subject-Object Swap &  \footnotesize{\texttt{\makecell{lexical\_overlap\_\\ln\_subject-object\_swap}}} & SO-Swap & \makecell{The doctor advised the president. $\not\rightarrow$\\The president advised the doctor.}\\\\ 
    \quad Preposition & \footnotesize{\texttt{\makecell{lexical\_overlap\_\\ln\_preposition}}} & Prep & \makecell{The tourist by the manager saw the artists.\\$\not\rightarrow$The artists saw the manager.}\\ 
    \textbf{\textit{Constituent (C)}} & & & \\
    \quad Embedded under if & \footnotesize{\texttt{\makecell{constituent\_\\cn\_embedded\_under\_if}}} & Embed-If & \makecell{If the artist slept, the actor ran.\\$\not\rightarrow$ The actor ran.}\\\\
    \quad Embedded under verb & \footnotesize{\texttt{\makecell{constituent\_\\cn\_embedded\_under\_verb}}} & Embed-Verb & \makecell{The lawyers believed that the tourists.\\shouted $\not\rightarrow$ The tourists shouted.}\\
    \bottomrule
    \end{tabular}
    \caption{Example cases and adversarial subcases from HANS \cite{bertsofafeather}. We focus mostly on these subcases in the main text as the model generalizes best to them. We use abbreviated names for ease of discussion.}
    \label{tab:hansexamples}
\end{table*}
\begin{table*}[t]
    \centering
    \small
    \begin{tabular}{ccc}
    \toprule
    \textbf{Question \#1} & \textbf{Question \#2} & \textbf{Equivalent?} \\
    \midrule
    Is a contagious yawn also fake? & Is a fake yawn also contagious? & \ding{55}\\
    How are noble gases stable? & How are stable gases noble? & \ding{55}\\
    How is Perth better than Melbourne? & How is Melbourne better than Perth? & \ding{55}\\
    
    \bottomrule
    \end{tabular}
    \caption{Example questions from the PAWS-QQP dataset~\cite{paws} dataset. We only work with questions whose label is ``not equivalent'', as the others are heuristic-friendly.}
    \label{tab:pawsqqpexamples}
\end{table*}

First, we check if these behaviorally different subnetworks consist of disjoint subsets of model components (attention heads).
To the contrary, we find a small set of nine attention heads that occur in \emph{all} subnetworks---even the ones that do not generalize at all.
Furthermore, we find that this same set of attention heads also appears when we prune earlier checkpoints of the model before it starts to generalize.
In the grokking setting, generalization is marked by a decrease in \emph{effective size}---defined as the size of the smallest subnetwork that matches the full network's performance on the generalization sets---as the model switches from a dense, memorizing subnetwork to a sparse, generalizing subnetwork.
In contrast, we find that generalization in our case is accompanied by a sharp \emph{increase} in effective size.
Together with the set of common attention heads, our results suggest that, instead of selecting between competing subnetworks, the model initially learns a ``core'' of attention heads that implement simple heuristics. 
The model ultimately generalizes by learning additional attention heads that interact with these heads.

We refer to this core set of attention heads as the \ti{heuristic core}, and conduct further analysis to better understand what roles they play in the model.
We find that these attention heads are associated with entailment heuristics---namely, attending to words that are repeated across sentences---supporting our characterization of these components as heuristic core.
On the other hand, ablating these attention heads from the original model leads to a dramatic \ti{decrease} in performance on the generalization set.
This result further supports the idea that generalization arises from additional components building off simple features extracted by heuristic components.
At intermediate sparsity\footnote{In this paper, sparsity refers to the fraction (or percentage) of attention heads and MLPs that have been pruned away.} levels, subnetworks contain different ``counter-heuristic'' components, leading to different degrees of partial generalization to different subcases.

These results have important practical implications. 
For instance, they suggest that we cannot make the model more robust by ablating the ``heuristic'' components.
More broadly, our experiments paint a more detailed picture of the mechanisms underlying generalization in natural language understanding tasks.
In doing so, they highlight interesting phenomena and open new avenues to future research into language models' internals.
\section{Problem Setup}
\label{sec:background}
We focus on two sentence-pair classification tasks.
MNLI~\citep{williams2018broad} is a natural language inference (NLI) dataset, which involves predicting whether a \ti{premise} sentence logically entails a \ti{hypothesis} sentence.
QQP~\citep{iyer2017first} is a paraphrase identification dataset.
Prior work has noted that these datasets admit ``heuristics'': solutions that achieve high in-domain accuracy but do not generalize well.
In particular,~\citet{mccoy2019right} introduced the HANS dataset, which tests whether NLI models use specific heuristics, such as lexical overlap (predicting that sentences entail each other if they share many words); see Appendix Table~\ref{tab:hansexamples} for examples and the list of abbreviations used for HANS subcases. 
Each subcase of HANS has $1000$ validation examples.
Similarly, the QQP-PAWS dataset~\citep{paws} tests whether paraphrase identification models are susceptible to a word-overlap heuristic (Appendix Table~\ref{tab:pawsqqpexamples}).
PAWS-QQP has $12,663$ examples in all, of which $8,696$ bear the verdict ``not equivalent''. We work with the latter subset in this paper.

\paragraph{Models.}
We mainly focus on BERT~\citep{bert} throughout the paper, but our findings also generalize to RoBERTa~\citep{liu2019roberta} and GPT-2~\citep{radford2019language} (Section~\ref{sec:roberta_gpt2}).
We fine-tune BERT models on MNLI and QQP 
(see Appendix~\ref{sec:pruneparams} for more training details).
Our results reproduce the observations of~\citet{smallexamplesgen}: in-domain accuracy saturates early, and out-of-domain accuracy increases much later.
We focus on the four subcases in Table~\ref{tab:hansexamples} for HANS, as the model generalizes well on them. Other subcases are discussed in Appendix~\ref{sec:fullresults}.
The effect is smaller, but still consistent, on PAWS-QQP (Table~\ref{tab:pawsqqpexamples}).

\begin{figure*}[t]
\centering
\subfloat[MNLI, $50\%$]{
    \includegraphics[width=0.34\linewidth]{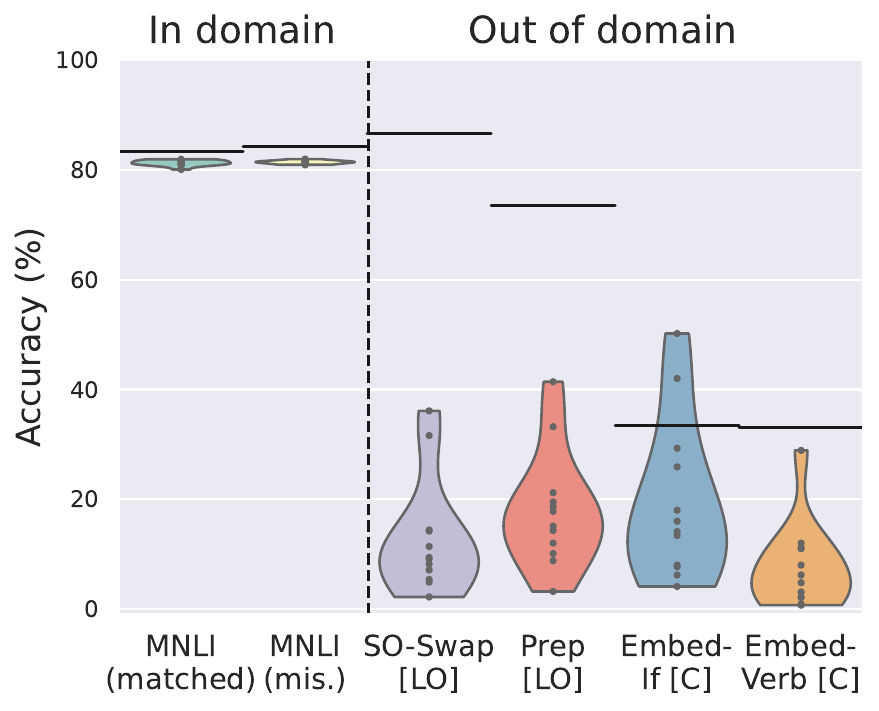}
    \label{fig:multiprune-a}
}
\subfloat[MNLI, $70\%$]{
    \includegraphics[width=0.305\linewidth]{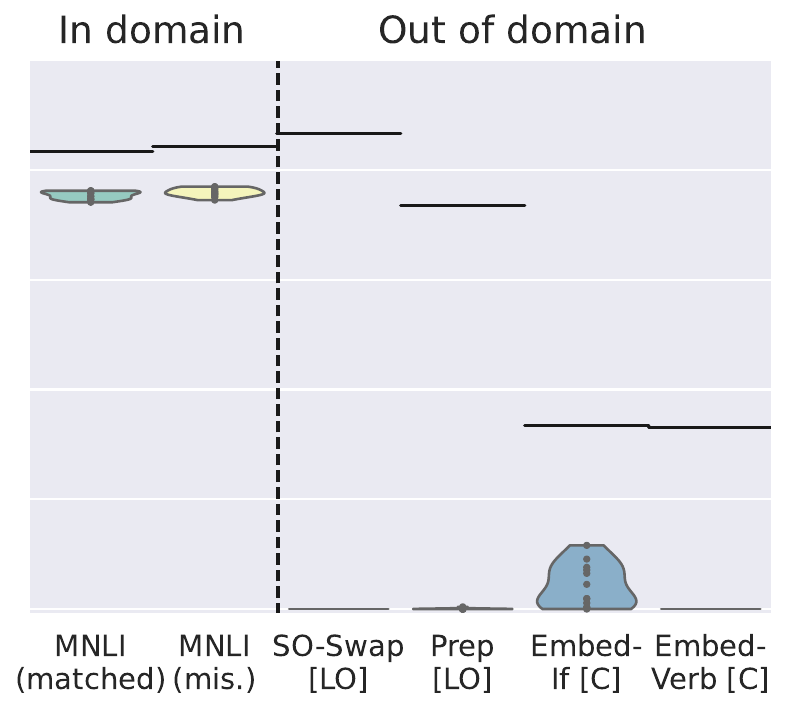}
    \label{fig:multiprune-b}
}
\subfloat[QQP, $30\%$]{
    \includegraphics[width=0.164\linewidth]{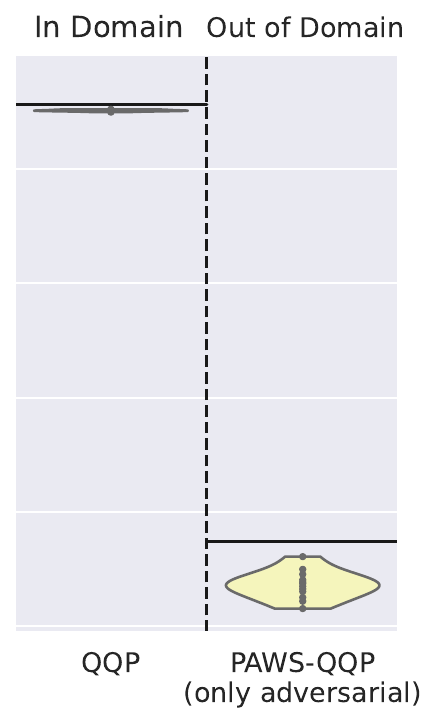}
    \label{fig:multiprune-c}
}
\subfloat[QQP, $60\%$]{
    \includegraphics[width=0.164\linewidth]{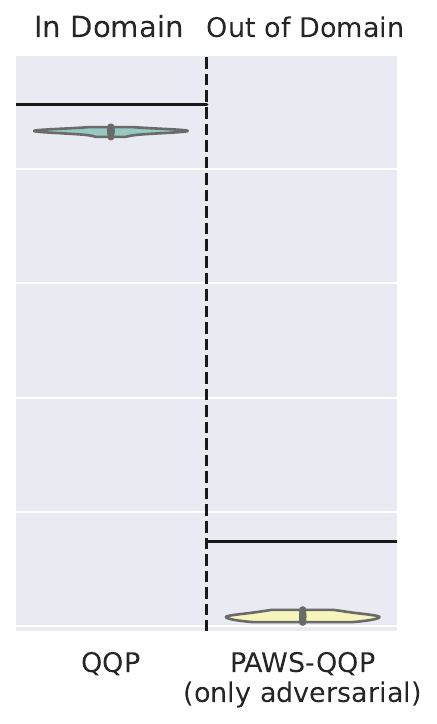}
    \label{fig:multiprune-d}
}
\caption{Pruning a BERT model with different random seeds results in subnetworks that perform similarly in-domain but generalize differently. The dots refer to the accuracy of the pruned subnetworks, while solid lines indicate full model performance. 
MNLI/HANS: At $50\%$ sparsity, the subnetworks perform within $3\%$ of the model on MNLI but show varying generalization. At $70\%$ sparsity, the subnetworks behave as pure heuristics despite respectable MNLI accuracy. The trend also holds for QQP/PAWS, with sparsities of $30\%$ and $60\%$. Figure~\ref{fig:multiprunefull} in Appendix~\ref{sec:fullresults} shows the plot for all subcases of HANS. 
}
\label{fig:multiprune}
\end{figure*}
\begin{figure*}
    \centering
    \includegraphics[width=\linewidth]{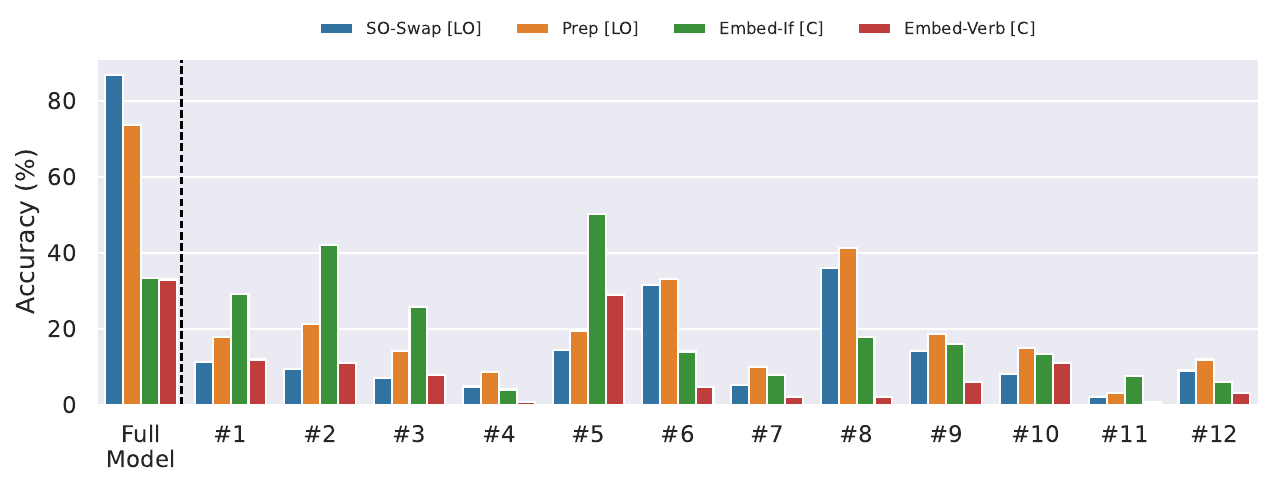}
    \caption{Different subnetworks at $50\%$ sparsity (found by pruning with different random seeds) generalize partially to different subcases of HANS. Subnetwork \#5 generalizes to the \emph{Constituent} subcases, whereas \#8 generalizes to the \emph{Lexical Overlap} subcases. Subnetwork \#2 does well on the \emph{Embed-If} subcase of \emph{Constituent}, but not \emph{Embed-Verb}.
    }
    \label{fig:diffsubdiffgen}
\end{figure*}

\paragraph{Finding subnetworks via pruning.}
For a Transformer model, we define a subnetwork as a subset of the attention heads and MLP (feed-forward) layers.
An important motif in our analysis will be the search for subnetworks that preserve model performance.
We choose structured pruning towards this end.
While circuit-finding methods~\citep{acdc} would also be useful, pruning is faster and allows injecting randomness more readily.
We use structured rather than unstructured pruning for two reasons.
First, unstructured pruning can be expressive enough to find subnetworks with completely different behavior from the original model~\citep{wen2023transformers}.
Second, structured pruning lets us compare subnetworks in terms of semantically meaningful components: attention heads.

We prune via optimization of a binary \emph{mask}, where $0$ and $1$ indicate dropping and retention, respectively.
Our runs prune attention heads and entire MLP layers, each corresponding to one bit in this mask.
We use \emph{mean ablation} per recommendations from work in Mechanistic Interpretability~\cite{bestpracticesactpatch}---pruned layers' (or heads') activations are replaced by the mean activation over the training data. 
Following CoFi Pruning \cite{cofi}, we relax binary masks to floats in $[0,1]$ and then perform gradient descent to optimize them. 
The objective optimized is the KL loss between the predictions of the full model and those of the pruned subnetwork.
Finally, the mask entries are discretized to $\{0,1\}$ based on a threshold. 
L0 regularization is used to enforce a target sparsity. 
We adopt \citet{l0kingma}'s recipe to model the masks and furnish the details in Appendix~\ref{sec:l0}. 
More details and hyperparameters, along with a more detailed discussion of the method, are provided in Appendix~\ref{sec:pruneparams}.
We freeze the model after fine-tuning and only optimize the pruning masks to preserve faithfulness to the full model.
\begin{figure*}[t]
\centering
\includegraphics[width=0.2\linewidth]{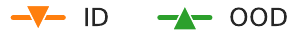}\\
\vspace{-1em}
\subfloat[MNLI/HANS]{
    \includegraphics[width=0.5\linewidth]{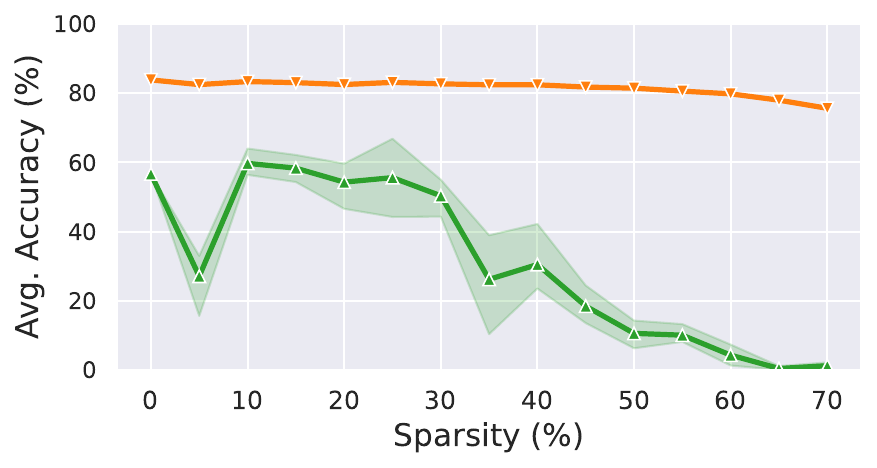}
}
\hfill
\subfloat[QQP/PAWS-QQP]{
    \includegraphics[width=0.47\linewidth]{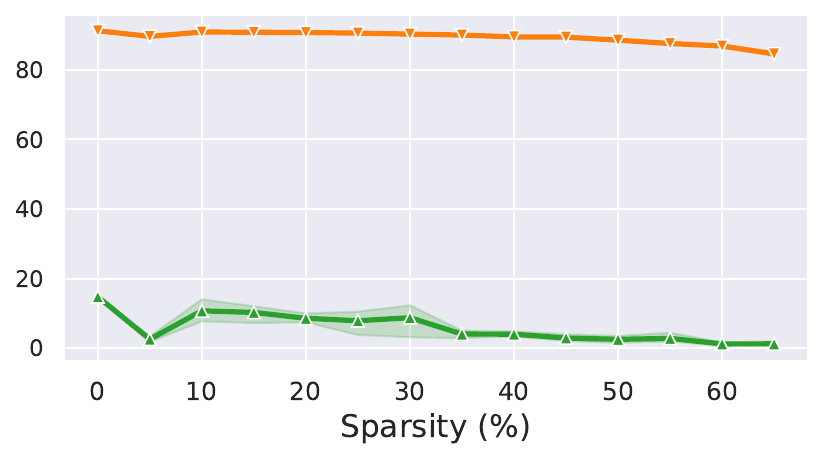}
}
\caption{The OOD accuracy decreases fairly smoothly with sparsity ($3$ seeds). The drop in ID accuracy is slow and has low variance. We find no subnetworks sparser than $30\%$ generalizing as well as the full model on either dataset.}
\label{fig:perfvssparsity}
\end{figure*}
\begin{figure}[t]
    \centering
    \includegraphics[width=\linewidth]{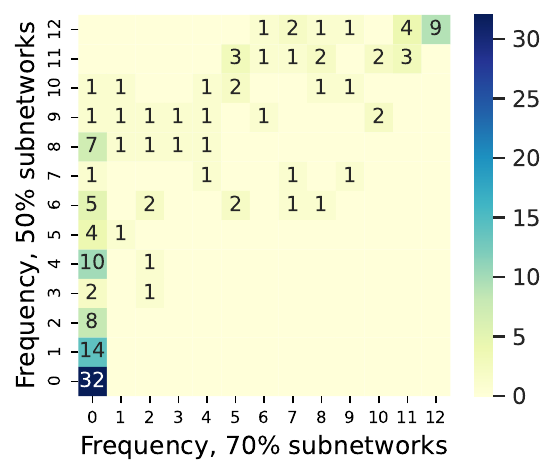}
    \caption{This heatmap quantifies the frequencies of attention heads in the $50\%$ and $70\%$ sparsity subnetworks (MNLI). Entry $(i,j)$ corresponds to the number of heads appearing in $i$/12 $50\%$ subnetworks and $j$/12 $70\%$ subnetworks. 
    In particular, we note that \ti{nine} attention heads appear in all of the subnetworks.
    }
    \label{fig:heatmap-5070}
\end{figure}
\section{Comparing Subnetwork Generalization}
We start by examining how different subnetworks generalize on MNLI/HANS and PAWS/QQP.
Specifically, we compare the generalization behavior of subnetworks with the same sparsity level, but pruned with different seeds; and we compare subnetworks pruned with different sparsity levels.
We prune the BERT models fine-tuned on MNLI and QQP with $12$ random seeds for different target sparsities (more details in Appendix~\ref{sec:pruneparams}).
The resulting subnetworks are evaluated on the ID (MNLI and QQP validation sets) and OOD (HANS subcases, and PAWS-QQP's adversarial subset) evaluation sets.
In our runs, we observe that the MLP layers are never pruned (although they are allowed to be).

\paragraph{Different subnetworks generalize differently.}
Figures~\ref{fig:multiprune-a} and~\ref{fig:multiprune-c} plot the results for a target sparsity of $50\%$ and $30\%$ in the case of MNLI and QQP, respectively.
The pruned models perform close to each other in-domain, and are all within 2-3\% of the full model's accuracy on both the MNLI and QQP validation splits.
However, their accuracies on the out-of-domain evaluation splits varies widely. 
In the most extreme case, the accuracies on the \emph{Embed-If (Constituent)} subcase of HANS vary from just $4.1\%$ to $50.2\%$. 
Moreover, different subnetworks generalize to different subcases, as Figure~\ref{fig:diffsubdiffgen} illustrates.
Subnetwork \#8 achieves $41.4\%$ at \emph{SO-Swap (Lexical Overlap)} but only $18.0\%$ at \textit{Embed-If (Constituent)}.
Subnetwork \#5, on the other hand, achieves only $19.5\%$ at the former but $50.2\%$ at the latter.
Thus, pruning different parts of the model seems to sacrifice generalization on the OOD subcases to varying degrees.
The results in the case of QQP and PAWS-QQP are similar.

\begin{figure*}[t]
    \centering
    \includegraphics[width=0.3\linewidth]{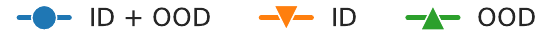}\\
    \vspace*{-1em}
     \subfloat[MNLI/HANS -- Accuracy]{
        \includegraphics[width=0.485\linewidth]{figures/eval\_all\_new.pdf}
    }
    \hfill
    \subfloat[QQP/PAWS-QQP (adv.) -- Accuracy]{
        \includegraphics[width=0.49\linewidth]{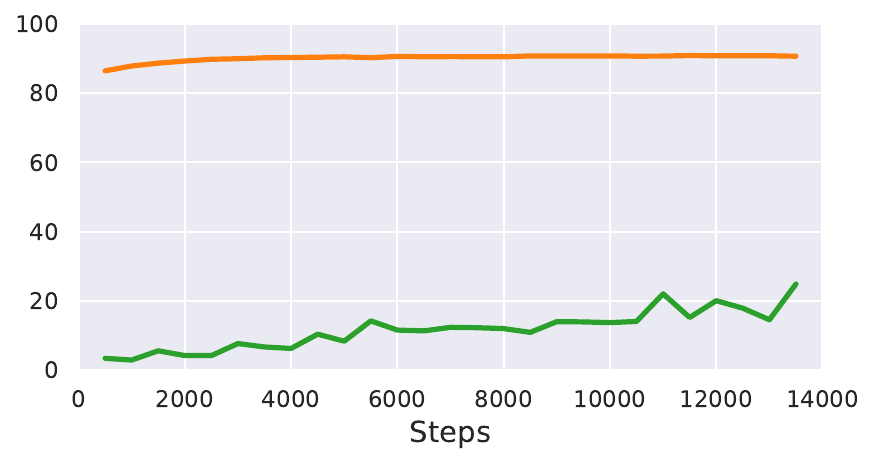}
    }\\
    \subfloat[MNLI/HANS -- Effective size]{
        \includegraphics[width=0.495\linewidth]{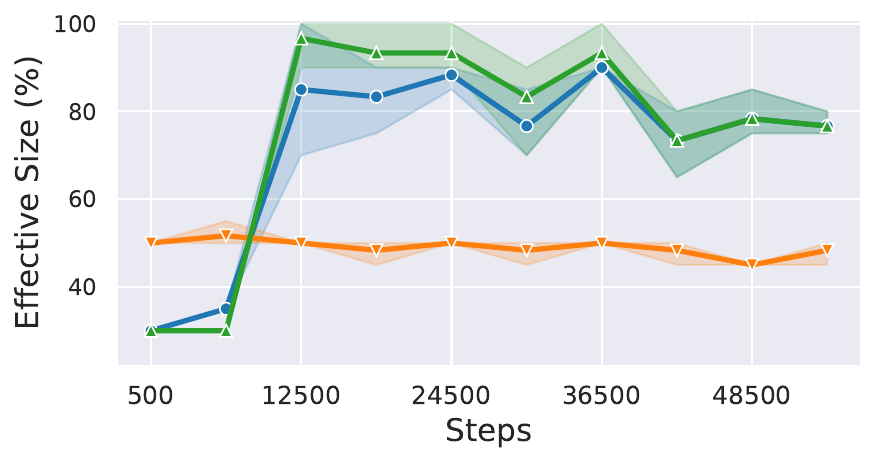}
    }
    \hfill
    \subfloat[QQP/PAWS-QQP (adv.) -- Effective size]{
        \includegraphics[width=0.48\linewidth]{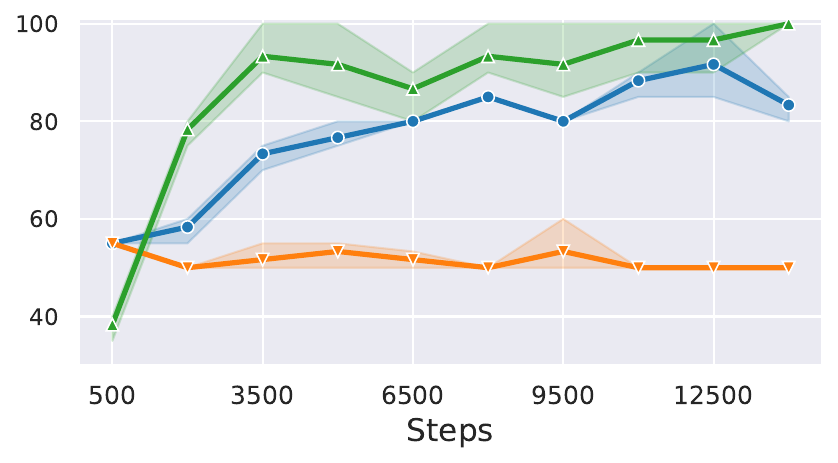}
    }\\
    \caption{The ID and OOD accuracies, and effective size of the model during fine-tuning, computed by pruning with $3$ seeds. Generalization is associated with a steep \emph{increase} in effective size for OOD/mixed datasets. The effective size for ID datasets remains constant, showing that these extra effective parameters go towards generalization. 
    }
    \label{fig:effsize}
\end{figure*}

\paragraph{Sparser subnetworks generalize worse.}
We also observe that sparser subnetworks consistently generalize worse.
First, we prune the model to a higher sparsity ($70\%$ for MNLI and $60\%$ for QQP) with $12$ random seeds.
As Figures~\ref{fig:multiprune-b} and~\ref{fig:multiprune-d} show, virtually \emph{all} subnetworks at this sparsity show a complete lack of generalization on almost every OOD subcase.
The ID accuracy stays at a respectable 74-76\%, indicating that a large portion of model performance is explained by these sparse subnetworks.
In Figure~\ref{fig:perfvssparsity}, we plot the generalization accuracy of subnetworks pruned at different sparsity levels, for $3$ random seeds.
The generalization accuracy generally drops steadily at higher sparsity levels and is devoid of any drastic phase transitions.

\section{Competing Subnetworks Hypothesis}
In the previous section, we observed that different subnetworks perform similarly to the full model in-domain but generalize differently.
Moreover, no sparse subnetwork generalizes as well as the full model.
We now seek to understand how these subnetworks emerge during training and interact to give rise to generalization.

\label{sub:csh}
Specifically, we explore if our findings can be explained in terms of \emph{competing subnetworks}: the model initially represents multiple solutions as different subnetworks, and it ultimately converges to one.
Whether it generalizes is dependent on which of the subnetworks it converges to.
This hypothesis has been found to explain the curious phenomenon of grokking~\citep{taleoftwocircuits}.
Additionally, as~\citet{smallexamplesgen} note, the ID performance of our BERT models saturates early in training (Figure~\ref{fig:effsize}) but the OOD accuracy starts to rise much later.
Motivated by the similarity to grokking, we test if generalization in our case also arises from competition between disjoint subnetworks.

We consider the Competing Subnetwork Hypothesis to consist of two main predictions, mirroring ~\citet{taleoftwocircuits}.
First, the model can be composed into subnetworks that are \ti{disjoint}, representing distinct algorithms consistent with the training data.
Second, a rise in generalization is accompanied by a reduction in the model's \ti{effective size}---the smallest subnetwork whose accuracy matches the model's within a given threshold (e.g., $3\%$) on a target dataset---which occurs when the model switches from a dense, non-generalizing subnetwork to a sparse, generalizing subnetwork.

\paragraph{All subnetworks share a common set of components.}
We start by measuring how many model components are shared between different subnetworks.
In particular, we compute the frequencies of each attention head in the $50\%$ (partially generalizing) subnetworks and in the $70\%$ (non-generalizing) subnetworks of the MNLI model.
These frequencies are visualized in Figure~\ref{fig:heatmap-5070} (a corresponding plot for QQP is presented in Figure~\ref{fig:heatmap-qqp-3060} of Appendix~\ref{sec:fullresults}).
The Spearman's Rho correlation between the two frequencies if $0.82$ ($p$-value = $1.6 \cdot 10^{-36}$), i.e., very strong agreement.
(The corresponding value for QQP is $0.74$, corresponding to strong agreement; see Figure~\ref{fig:membership-qqp} in Appendix~\ref{sec:fullresults}.)
In particular, we observe that \tf{nine attention heads} appear in \emph{all} subnetworks, even the sparsest ones---that is, the subnetworks that behave most consistently with the heuristics.
In other words, instead of finding disjoint subnetworks, we find a high degree of overlap (the expected intersection of $12$ random $70\%$ sparsity subnetworks is $7.7 \cdot 10^{-5}$ heads) between the subnetworks that partially generalize and the subnetworks that do not generalize at all.

\begin{figure*}[t]
\centering
\subfloat[Layer $2$, Head $5$]{
    \includegraphics[width=0.24\linewidth]{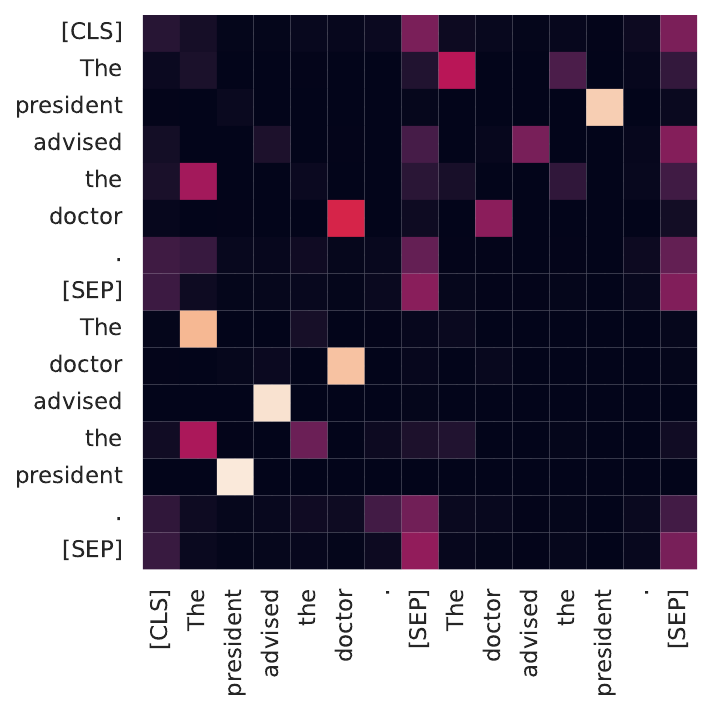}
}
\subfloat[Layer $3$, Head $7$]{
    \includegraphics[width=0.24\linewidth]{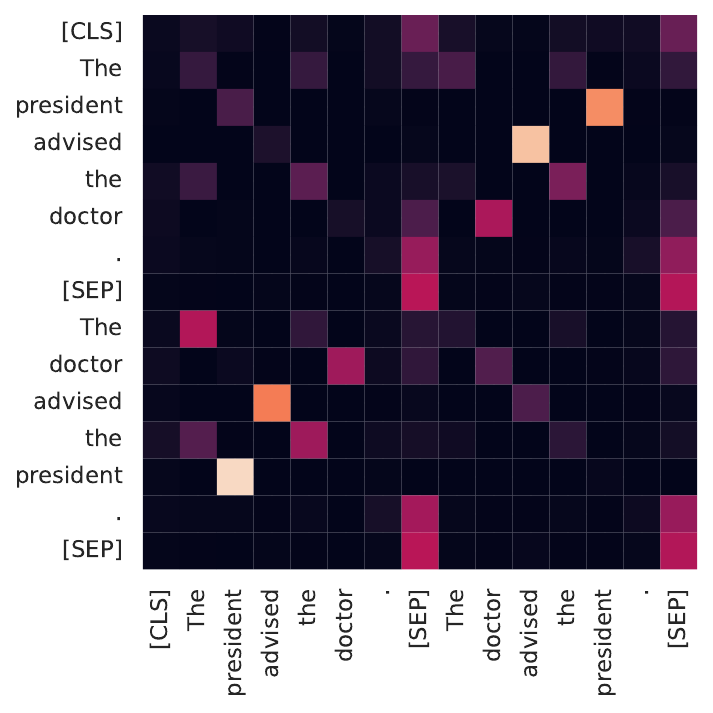}
}
\subfloat[Layer $4$, Head $0$]{
    \includegraphics[width=0.24\linewidth]{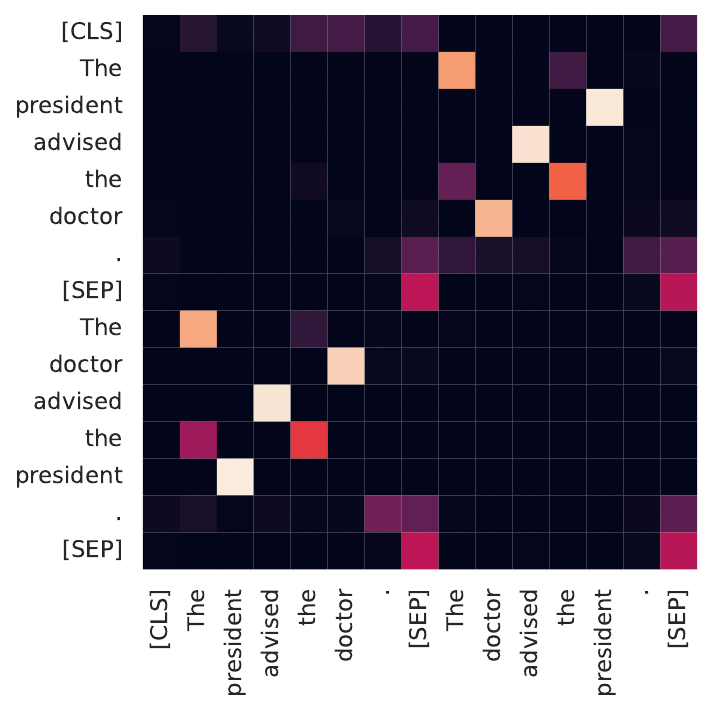}
}
\subfloat[Layer $8$, Head $11$]{
    \includegraphics[width=0.28\linewidth]{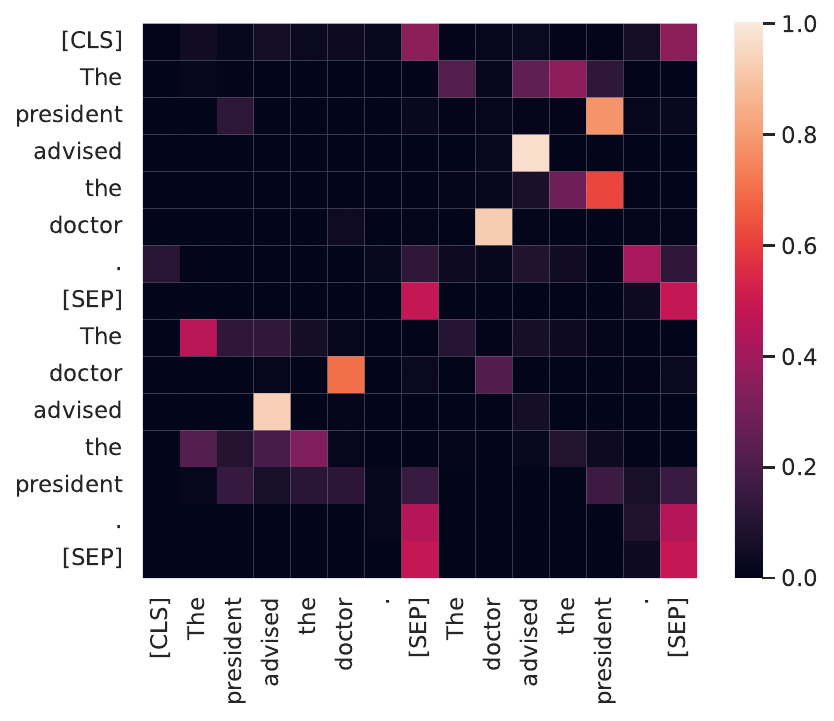}
}
\caption{Many attention heads common to the MNLI subnetworks attend to tokens co-occurring in the premise and context. Since the order of, \emph{e.g.}, ``doctor'' and ``president'' is reversed, the heads seem to identify lexical overlap.}
\label{fig:attnmaps}
\end{figure*}
\begin{figure*}[t]
\centering
\subfloat[Layer $2$, Head $5$]{
    \includegraphics[width=0.24\linewidth]{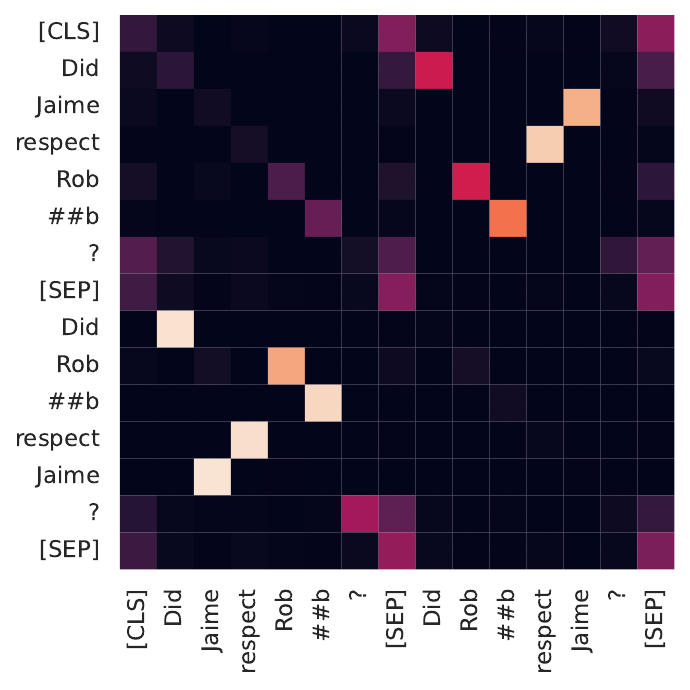}
}
\subfloat[Layer $4$, Head $0$]{
    \includegraphics[width=0.24\linewidth]{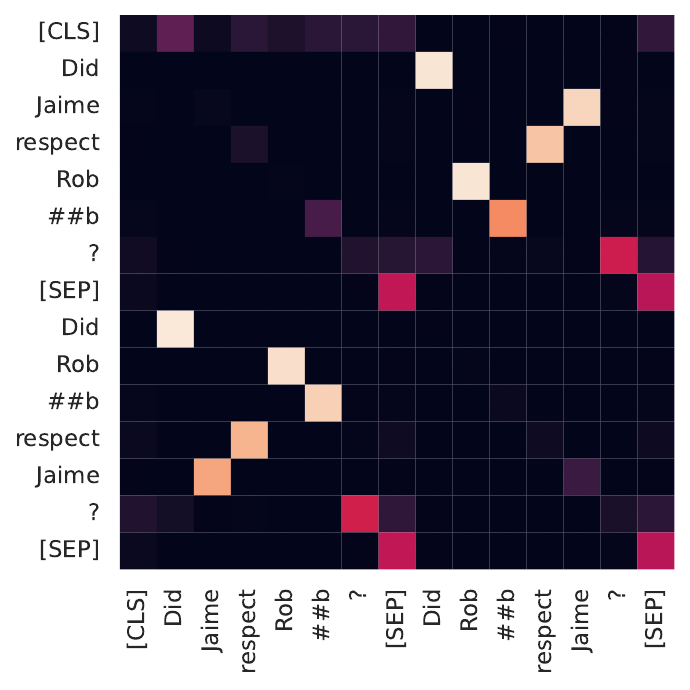}
}
\subfloat[Layer $8$, Head $11$]{
    \includegraphics[width=0.24\linewidth]{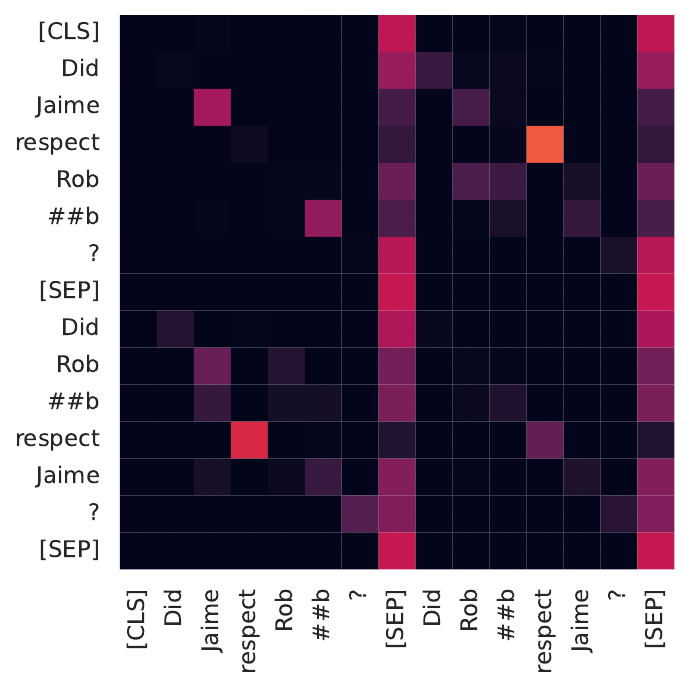}
}
\subfloat[Layer $9$, Head $1$]{
    \includegraphics[width=0.28\linewidth]{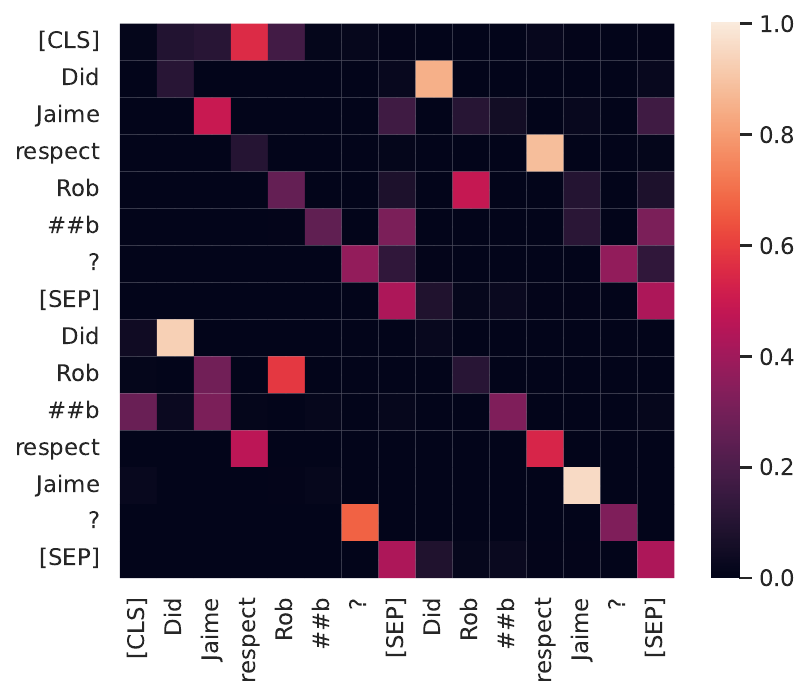}
}
\caption{Many attention heads in the QQP heuristic core also focus their attention between tokens repeated across the premise and context. Interestingly, there is a significant overlap between these heads and those for MNLI.}
\label{fig:attnmaps-qqp}
\end{figure*}
\begin{figure}[t]
    \centering
    \includegraphics[width=\linewidth]{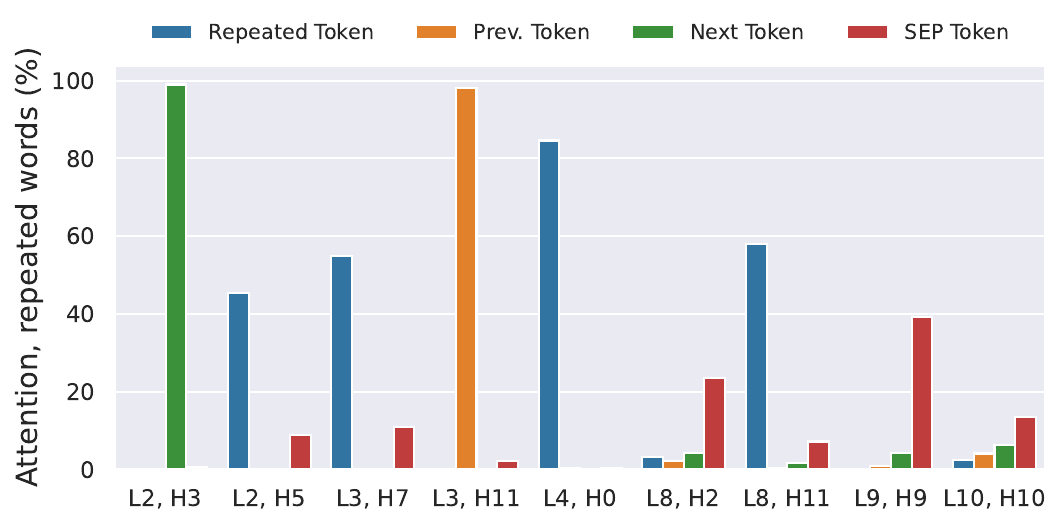}
    \caption{Of the $9$ heads occurring in all MNLI subnetworks, several assign their attention weight between words repeated across the premise and hypothesis. Others either attend to the previous, next, or separator token. Here, ``L$x$, H$y$'' stands for ``Head $y$ of Layer $x$''. 
    }
    \label{fig:attention}
\end{figure}

\paragraph{Effective size increases as the model generalizes.}
\begin{table*}
    \centering
    \small
    \begin{tabular}{ll|llll}
        \toprule
        \multirow{2.5}{*}{\textbf{\makecell{Ablated Head}}} & \multicolumn{5}{c}{\textbf{Accuracy (\%)} 
 $\uparrow$}\\
        \cmidrule{2-6}
         & \textbf{\makecell{MNLI (matched)}} & 
\textbf{\makecell{SO-Swap [LO]}} & \textbf{\makecell{Prep [LO]}} & \textbf{\makecell{Embed-If [C]}} & 
\textbf{\makecell{Embed-Verb [C]}}\\
        \midrule
        None & 83.3 & 86.7 & 73.5 & 33.0 & 33.4\\
        Avg. (Non-HC) & 83.2$_{\downarrow{0.1}}$ & 86.2$_\downarrow{0.5}$ & 73.1$_{\downarrow{0.4}}$ & 33.5$_{\uparrow{0.5}}$ & 34.1$_{\uparrow{0.7}}$\\
        \midrule
        Layer 2, Head 3 & 83.2$_{\downarrow 0.1}$ & 79.3$_{\redone{7.4}}$ & 72.3$_{\downarrow 1.2}$ & 28.8$_{\redone{4.2}}$ & 28.9$_{\redone{4.5}}$\\
        Layer 2, Head 5 & 83.4$_{\uparrow 0.1}$ & 70.3$_{\redtwo{16.4}}$ & 58.6$_{\redtwo{14.9}}$ & 28.5$_{\redone{4.5}}$ & 23.6$_{\redone{9.8}}$\\
        Layer 3, Head 7 & 83.5$_{\uparrow 0.2}$ & 68.6$_{\redtwo{18.1}}$ & 60.2$_{\redtwo{13.3}}$ & 28.2$_{\redone{4.8}}$ & 23.0$_{\redtwo{10.4}}$\\
        Layer 3, Head 11 & 83.5$_{\uparrow 0.2}$ & 81.8$_{\redone{4.9}}$ & 70.1$_{\downarrow 3.4}$ & 31.0$_{\downarrow 2.0}$ & 32.9$_{\downarrow 0.5}$\\
        Layer 4, Head 0 & 82.9$_{\downarrow 0.4}$ & 45.4$_{\redthree{41.3}}$ & 41.8$_{\redthree{31.7}}$ & 21.2$_{\redtwo{11.8}}$ & 12.7$_{\redthree{20.7}}$\\
        Layer 8, Head 2 & 83.7$_{\uparrow 0.4}$ & 84.5$_{\downarrow 2.2}$ & 69.5$_{\downarrow 4.0}$ & 31.6$_{\downarrow 1.4}$ & 27.5$_{\redone{5.9}}$\\
        Layer 8, Head 11 & 81.0$_{\downarrow 2.3}$ & 88.3$_{\uparrow 1.6}$ & 76.5$_{\uparrow 3.0}$ & 37.6$_{\blueone{4.6}}$ & 34.1$_{\uparrow 0.7}$\\
        Layer 9, Head 9 & 83.4$_{\uparrow 0.1}$ & 87.7$_{\uparrow 1.0}$ & 74.2$_{\uparrow 0.7}$ & 36.1$_{\uparrow 3.1}$ & 32.4$_{\downarrow 1.0}$\\
        Layer 10, Head 10 & 81.5$_{\downarrow 1.8}$ & 84.7$_{\downarrow 2.0}$ & 72.2$_{\downarrow 1.3}$ & 41.1$_{\bluetwo{8.1}}$ & 38.7$_{\bluetwo{5.4}}$\\
        \midrule
        \makecell{L2.H5 \& L3.H7 \& L4.H0} & 80.4$_{\downarrow 2.9}$ & 3.6$_{\redultra{83.1}}$ & 11.1$_{\redultra{62.4}}$ & 20.1$_{\redtwo{12.9}}$ & 4.7$_{\redthree{28.7}}$\\
        \bottomrule
    \end{tabular}
    \caption{The effects of ablating the $9$ heads occurring in all MNLI subnetworks individually from the full model. It is noteworthy that some of these are also important for counter-heuristic behavior, as their ablation leads to reduced generalization. In the final row, L$x$.H$y$ refers to Head $y$ in Layer $x$. 
    }
    \label{tab:ablations}
\end{table*}
\begin{table}[t]
    \centering
    \small
    \begin{tabular}{lr|r}
        \toprule
        \multirow{2.5}{*}{\textbf{\makecell{Ablated Head}}} & \multicolumn{2}{c}{\textbf{Accuracy (\%)} 
 $\uparrow$}\\
        \cmidrule{2-3}
         & \textbf{\makecell{QQP}} & 
\textbf{\makecell{QQP-PAWS (adv.)}}\\
        \midrule
        None & 91.2 & 14.8\\
        \midrule
        Layer 2, Head 3 & 91.1$_{\downarrow 0.1}$ & 14.1$_{\downarrow 0.7}$\\
        Layer 2, Head 5 & 91.1$_{\downarrow 0.1}$ & 9.9$_{\redtwo{4.9}}$\\
        Layer 3, Head 3 & 91.2$_{\downarrow 0.0}$ & 13.9$_{\downarrow 0.9}$\\
        Layer 4, Head 0 & 90.2$_{\downarrow 1.0}$ & 3.3$_{\redthree{11.6}}$\\
        Layer 8, Head 11 & 91.0$_{\downarrow 0.2}$ & 14.6$_{\downarrow 0.2}$\\
        Layer 9, Head 1 & 91.2$_{\downarrow 0.0}$ & 16.6$_{\blueone{1.8}}$\\
        Layer 9, Head 9 & 91.1$_{\downarrow 0.1}$ & 16.2$_{\blueone{1.4}}$\\
        Layer 11, Head 6 & 91.1$_{\downarrow 0.1}$ & 17.1$_{\bluetwo{2.3}}$\\
        \bottomrule
    \end{tabular}
    \caption{The effects of ablating the $8$ heads occurring in all QQP subnetworks individually from the full model. The heads with the strongest reduction are the same ones that had the largest effect on the MNLI model.}
    \label{tab:ablations-qqp}
\end{table}
Next, we investigate how these subnetworks emerge over training.
We conduct a similar analysis to~\citet{taleoftwocircuits} and plot the \ti{effective size} of the model, defined as the size of the smallest subnetwork that is \ti{faithful} to the model on a particular evaluation dataset $D$.
We say that a subnetwork is \ti{faithful} to a model on dataset $D$ if their accuracies are within $3\%$.
We report faithfulness using different choices of $D$---either the in-domain validation set, the out-of-domain validation sets, or the equally weighted combination of the two (mixed data).
We prune each checkpoint to sparsities of multiples of $5\%$ and pick the highest one that yields a faithful subnetwork.
The reported accuracies at each sparsity are averages over three seeds.

The results are in Figure~\ref{fig:effsize}.
The effective size needed to approximate the model on the in-domain evaluation remains relatively flat over the course of training.
However, on out-of-domain and mixed data, generalization is accompanied by an \ti{increase} in effective size.
We also examine the set of attention heads that appear in the MNLI subnetworks at the checkpoint immediately before the model starts to generalize.
We find that the same nine attention heads that appear in all subnetworks at the end of training appear in these subnetworks as well.
This suggests that, rather than switching between competing subnetworks, the model first learns a set of attention heads that compute simple, non-generalizing features, and then generalizes by incorporating more components into this core.

\section{Understanding the Heuristic Core}
Instead of finding evidence for distinct, competing subnetworks, we have found that all subnetworks share a set of common components.
In particular, nine attention heads appear in all MNLI subnetworks, including subnetworks that do not generalize at all.
We find the same nine attention heads in subnetworks when we prune checkpoints early in training, before the model generalizes.
We refer to these nine attention heads as the \ti{heuristic core} of the model.
In this section, we inspect these attention heads in more detail, to understand what features they compute, and how they interact with the rest of the model to enable generalization.

\paragraph{How do the heuristic core attention heads behave?}
First, we look at the attention patterns exhibited by these attention heads.
We inspect example attention patterns and observe that most heads exhibit a simple attention pattern, and we quantify these in Figure~\ref{fig:attention}.
Four out of nine heads (MNLI) attend to tokens that repeat across the premise and hypothesis, suggesting that they extract token overlap features.
Figure~\ref{fig:attnmaps} shows example attention patterns.
Notably, we find that eight attention heads also appear in all QQP subnetworks.
Figure~\ref{fig:attnmaps-qqp} presents an example indicating that these heads play a similar role for QQP, attending between repeated words.

Other low-layer heads attend to the previous or subsequent token.
The heads in the higher layers are more difficult to characterize but often attend to the special separator token, perhaps aggregating information from across the sequence.
Overall, these findings are consistent with the notion that these attention heads calculate simple, shallow features.

\paragraph{How do the heuristic core attention heads interact with the rest of the model?}
Next, we ablate each heuristic core attention head from the full model and observe how the in-domain and out-of-domain accuracies change, in Table~\ref{tab:ablations}.
None of the heads' ablation affects in-domain accuracy greatly.
Surprisingly, however, many heads seem to be critical to performing well on HANS:
for example, ablating a single head (L4.H0) reduces performance on \emph{SO-Swap} by more than 40\%.
Moreover, the drops observed on ablating these heads is \emph{super-additive}: ablating just three of them leads to the OOD accuracy plummeting. 
This indicates that heads outside the heuristic core rely on the heuristic core to implement more complex behavior.

Table~\ref{tab:ablations-qqp} paints a similar picture for QQP. 
Perhaps most interestingly, the heads $2.5$ and $4.0$ that had the maximum ablation effect for MNLI also stand out here.
Along with the fact that $5$ of these $8$ heads also exist in the MNLI model's core, this highlights the possibility that these heads already extract word overlap features in the BERT model prior to fine-tuning. We leave this question for future work.


\section{Findings Generalize to RoBERTa and GPT-2}
\label{sec:roberta_gpt2}

We show in Appendix~\ref{ap:roberta} that our results extend to RoBERTa~\citep{liu2019roberta} models.
Appendix~\ref{ap:gpt2} finds that they also transfer to GPT-2~\citep{radford2019language} models, albeit with some novel behavior.
Specifically, we observe that the accuracy on the \emph{non-adversarial} OOD (`entailment') cases slightly drops late into training, and is accompanied by a mild increase in effective size.
The performance on these cases also varies across different subnetworks of the same sparsity.
Explaining these additional features requires further research and makes for future work.
We also note that it is possible that models much larger (such as those with 7B+ parameters) take a different route to generalization.
They might thus show qualitatively different behavior, and we urge caution in applying the results of this paper to them.
\section{Related Work}
\label{sec:relatedwork}
\paragraph{Pruning.} Pruning \cite{pruning} removes layers or parameters from a network for space efficiency and speed-ups. 
Structured Pruning \cite{structuredpruning, llmpruner, cofi, shearedllama} preserves some structure in the model, e.g. equal fraction of surviving parameters in each attention head, accepting lower compression in exchange for higher speed-ups. 
Unstructured pruning \cite{wanda} approaches like Magnitude Pruning \cite{pruning, pruning2} achieve better compression at the cost of lower potential speedup. 
Proposals such as Layer dropping \cite{layerdropping} and Block Pruning \cite{blockpruning} have explored pruning at coarser granularities.

\paragraph{Pruning for interpretability.} Mechanistic Interpretability seeks to understand models through their components, e.g., neurons, attention heads, and MLPs. 
Such analysis uncovers intricate interactions between different atomic components, which form \emph{circuits} \cite{mech-interpretability, circuits1} implementing various tasks. 
Pruning has occasionally been used as part of a larger interpretability effort \cite{neurosurgeon}, though mostly in an ad-hoc manner. 
\citet{pruningonft} use pruning and probing techniques to show that fine-tuning learns a thin ``wrapper'' around model capabilities. 
\citet{structcompostpruning} use pruning to identify subnetworks which they then show form modular building blocks of model behavior. 
In a similar vein, \cite{inputmasking} learn differentiable masks over the input to investigate model behavior.
While most of these approaches replace missing layers with an identity function, work in Mechanistic Interpetability~\cite{bestpracticesactpatch, acdc} has shown that adding their mean activations might preserve faithfulness better.
Prior work has examined attention patterns to decipher model behavior~\citep[e.g. ][]{clark-etal-2019-bert}, although attention patterns have also been shown to be misleading in some cases~\citep{jain-wallace-2019-attention}.

\paragraph{Generalization.} The question of how a neural network generalizes has remained a compelling research direction for the NLP community \cite{generalization}. 
\citet{bertsofafeather} find that fine-tuning BERT with different random seeds leads to disparate generalization despite similar in-domain accuracies.
Other works have identified tasks where OOD accuracies increase long after in-domain accuracy saturates and attributed them to hard minority examples \cite{smallexamplesgen, hardtolearn}. 
Recently, a phenomenon called Grokking~\cite{grokking} was discovered, where test accuracy on a toy algorithmic task increases abruptly long after overfitting to the training data.
It has been explained in terms of competing subnetworks~\cite{taleoftwocircuits, grokkingexplaindeepmind}.
Grokking-like phenomena have also been identified in other settings \cite{omnigrok} and for small transformers~\cite{grokkingtransformers} on hierarchical depth generalization.
Simultaneously, questions about memorizing in pretrained LMs have been asked in other settings~\cite{memvsgen}.
\citet{friedman2023interpretability} document cases where simplifying a Transformer's representations preserves in-domain performance but impairs systematic generalization; we discover a similar ``generalization gap'' for another simplification method, pruning.

\paragraph{Subnetworks as modular components.} An emerging body of work \cite{structcompostpruning, modularitymultilingual, modularity1} suggests that neural networks and language models may consist of modular subnetworks each performing different subtasks. 
\citet{knowledgesubnetwork} find a subnetwork in GPT-2 responsible for most of its factual knowledge -- removing it does not lead to loss of performance on non-knowledge-intensive tasks. 
\citet{ablationbehavior} report the possibility of ablating away undesired behavior in a model (e.g. toxicity), while~\citet{wei2024assessing} identify regions that when pruned away compromise the safety alignment of a Language Model. 
In concurrent work,~\citet{zhang2024unveiling} find a subset of less than 1\% of an LM's parameters vital for linguistic competence across $30$ languages---which they call the ``Linguistic Core''.
\citet{prakash2024finetuning} report that entity-tracking circuits in fine-tuned models exist even before fine-tuning, providing a potential explanation for why the heuristic core heads we found for MNLI and QQP had a substantial overlap.~\citet{panigrahi2023a} find that grafting $\sim$0.01\% of model parameters from a fine-tuned version can recover most of the gains from fine-tuning.
\section{Conclusion}
\label{sec:conclusion}
We showed that pruning a pretrained LM with different random seeds can yield subnetworks that perform similarly to the original model in-domain but generalize differently.
Moreover, sparser subnetworks generalize worse.
To explain this result, we first considered the \ti{Competing Subnetworks Hypothesis}, which has been used to explain grokking on algorithmic tasks; it suggests that the model consists of disjoint subnetworks representing different solutions, some which generalize and some which do not.
We found that our results are instead more consistent with the existence of a \ti{Heuristic Core}: a subset of attention heads that appear in all subnetworks and calculate shallow, non-generalizing features. The model generalizes by incorporating additional heads that interact with the heuristic core.
Our results have practical implications about the effect of pruning on generalization and raise intriguing questions about how language models learn to generalize by composing simple components.

\section*{Limitations}
Our approaches are not without limitations.
Our analysis is based on a top-down approach of identifying subnetworks. 
Thus, we do not look for or discover ``circuits'' that are atomically interpretable.
Our experiments only use text classification tasks, although we believe that such a task is ideal due to clear definitions of heuristics.
Additionally, we only work with BERT models---it is possible that larger transformers, or those pretrained differently, exhibit different behavior.
Finally, although we have analyzed the mechanisms of generalization in this paper, the question of why they are undertaken remains elusive for the moment.
Our experiments also raised many novel questions, which require further research to answer.

\section*{Ethical Considerations}
We do not foresee any ethical considerations raised due to our work. 
We work with pretrained language models, which can at times hallucinate or demonstrate other undesirable behavior.
Our work, however, does not interact with these facets of model behavior.
Additionally, we only work with English datasets in our paper.

\section*{Acknowledgements}
We thank Michael Tang for his detailed suggestions and alacritous insights, without which this project would be incomplete.
We are grateful to Alexander Wettig and Zirui Wang for their feedback on paper drafts.
Finally, we thank everyone at the Princeton NLP group---especially Mengzhou Xia and Tianyu Gao, for fruitful discussions regarding our experiments.
This research is funded by the National Science Foundation (IIS-2211779), a Sloan Research Fellowship, and a Hisashi and Masae Kobayashi *67 Fellowship.


\bibliography{ref}

\newpage
\appendix

\section{Model and Training Details}
\label{sec:pruneparams}
\paragraph{BERT Architecture}
\label{app:model_details}
Our experiments use fine-tuned BERT models~\cite{bert}.
In the simplest setting, a Transformer Model \cite{transformers} consists of $L$ blocks, where each block is in turn composed of a Multi-Head \emph{Attention} layer followed by an \emph{MLP} layer. 
Given an input $X$, the Attention Layer outputs
$$X + \sum_{i=1}^{N_h} \text{Attn}(W_Q^{(i)}, W_K^{(i)}, W_V^{(i)}, X)W_O^{(i)}$$
where $N_h$ is the number of attention heads, and $W_K^{(i)}$, $W_Q^{(i)}$ and $W_V^{(i)}$ denote the Key, Query, and Value matrices, respectively. 
$W_O^{(i)}$ performs a linear output projection, before adding the result to the \emph{residual stream} via a skip connection. 
On the other hand, an MLP layer first up-projects its input $X$ by multiplying with $W_U \in \mathbb{R}^{d \times d_h}$, then passes the result through an activation function such as GeLU, then down-projects the result back via $W_D \in \mathbb{R}^{d \times d_h}$. 
Here $d$ is the hidden dimension and $d_h$ typically equals $4d$. 
This is then added to the residual stream to produce
$$X + \text{GeLU}(XW_U)W_D$$
Layer Normalization precedes both layers.

\paragraph{BERT Fine-Tuning} We fine-tuned BERT (\texttt{bert\_base\_cased}, 125M parameters) on MNLI for $5$ epochs ($61360$ steps with batch size $= 32$), with a learning rate of $2\cdot 10^{-5}$, a and a maximum sequence length of $128$. Checkpointing and evaluation were performed every $500$ steps. 
For QQP, we opted for $40000$ steps of fine-tuning with an effective batch size of $256$ and a learning rate of $2\cdot 10^{-5}$. Evaluation and checkpoint were once again performed every $500$ steps.
All runs used the Adam optimizer \cite{adam}, with $\beta = (0.9, 0.999)$ and $\epsilon = 10^{-8}$. 

Each run takes approximately 3 hours to complete on one NVIDIA A5000 GPU.
\section{Details of Pruning and L0 Regularization}
\label{sec:l0}
Our formulation of the masks is the same as CoFi Pruning \cite{cofi}'s, and is built on top of \citet{l0kingma}'s work. Specifically, the masks $z$ are modeled as hard concrete distributions:
$$\mathbf{u} \sim \text{Uniform}(\epsilon,1-\epsilon)$$
$$\textbf{s} = \sigma\left(\frac{1}{\beta} \cdot \frac{\mathbf{u}}{1-\mathbf{u}} + \log \mathbf{\alpha}\right)$$
$$\tilde{\mathbf{s}} = \mathbf{s} \times (r-l) + l$$
$$\mathbf{z} = \min(1, \max(0, \tilde{\mathbf{s}}))$$
Here, $\sigma$ denotes the sigmoid function, and $\epsilon = 10^{-6}$ is chosen to avoid division by $0$ errors. The temperature is fixed at $\frac{1}{\beta} = \frac{2}{3}$. The result is then stretched to the interval $[l,r] = [-0.1,1.1]$, with the probability mass from $[-0.1,0]$ and $[1,1.1]$ accumulated to $0$ and $1$ in the end. This ensures that the mask is incentivized to contain values close to $0$ or $1$. Here, the log alphas $\log \mathbf{\alpha}$ are the main learnable parameters.

Following \citet{structuredpruning} and \citet{cofi}, we enforce a target sparsity via a Lagrangian Term. Supposing the target and current sparsity to be $s$ and $t$, we add
$$\mathcal L_s = \lambda_1 (t-s) + \lambda_2 (t-s)^2$$
to the loss (CoFi uses $s-t$ instead of $t-s$, but the two are functionally the same; consider the transform $\lambda_1 \longrightarrow -\lambda_1$). Along with training the model parameters, a gradient \emph{ascent} is performed on $\lambda_1$ and $\lambda_2$, so that the updates of the lambdas always keep adjust the contribution of $\mathcal L_s$ to keep the sparsity in check.

\paragraph{Pruning - General Commentary} We found that the hyperparameters were very sensitive to the choice of learning rate.
Deviation even by a factor of $3$ would cause instability in training. 
Specifically, if the learning rate for the ascent on the Lagrangian Parameters ($\lambda_1$ and $\lambda_2$) was too low, the model would stay under a sparsity of $10^{-5}$ until the target sparsity climbed to $0.15$ or so, and would then jump to a value beyond the target, sacrificing a lot of performance in the process. 
On the other hand, a large value also leads to insufficient time for ``settling into'' the intermediate sparsities. 
The exact choice of the number of optimization steps was also important, although not as sensitive as the learning rate. 

\begin{figure}[t]
    \centering
    \includegraphics[width=\linewidth]{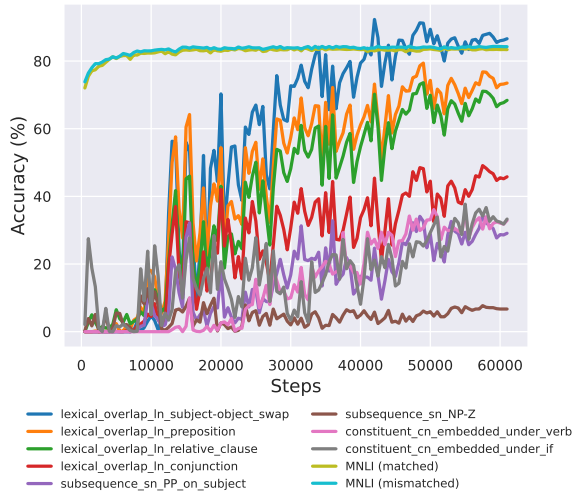}
    \caption{Model performance saturates early on MNLI, but generalizing traits are learned much later.}
    \label{fig:bertgrokfull}
\end{figure}

\begin{figure}
    \centering
    \includegraphics[width=\linewidth]{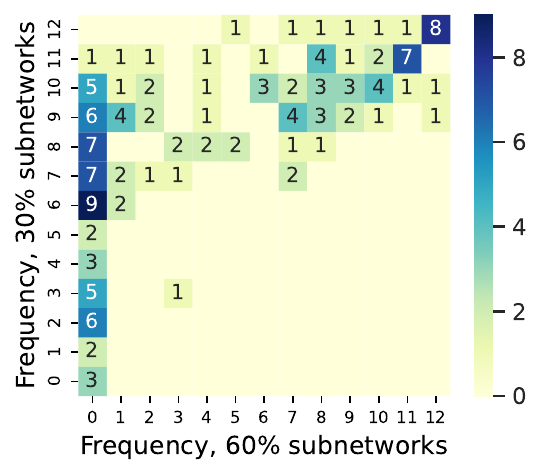}
    \caption{This heatmap quantifies the frequencies of attention heads in the $30\%$ and $60\%$ sparsity subnetworks of the QQP model. Entry $(i,j)$ denotes the number of heads appearing in $i/12$ $30\%$ subnetworks and $j/12$ $60\%$ subnetworks. Eight attention heads appear in all subnetworks.}
    \label{fig:heatmap-qqp-3060}
\end{figure}

\paragraph{BERT Pruning} We used a maximum sequence length of $128$ as before, with an effective batch size of $128$ including gradient accumulation. 
The learning rates for log alphas and lambdas were $0.1$ and $1$, respectively. 
We use the formulae
$$\texttt{warmup\_steps} = 6500 + 50 \times \texttt{sparsity\_pct}$$
$$\texttt{total\_steps} = \texttt{warmup\_steps} + 6 \times \texttt{sparsity\_pct}$$
to compute the number of steps taken to linearly warm up the target sparsity, and the total number of steps, respectively.
For sparsity targets of $50\%$, these translate to $9300$ training steps, of which the first $9000$ warm up the target sparsity. 
For $70\%$ target, these change to $10500$ and $10000$, respectively. 
Evaluation and checkpointing were done every $64$ steps but we always used the final checkpoint to report results. 
The thresholds were chosen with a grid search on $[0,1]$ with stride $0.05$ to match the target sparsity as closely as possible.
They were found to always be $0.5$ and $0.35$ for target sparsities of $0.5$ and $0.7$.

Each pruning run takes approximately 90 minutes to complete on an A5000 GPU.
We estimate our total computational budget to be 750 GPU hours.

\section{More results}
\label{sec:fullresults}
\begin{figure*}[t]
\centering
\subfloat[Sparsity $= 50\%$]{
    \includegraphics[width=0.5\linewidth]{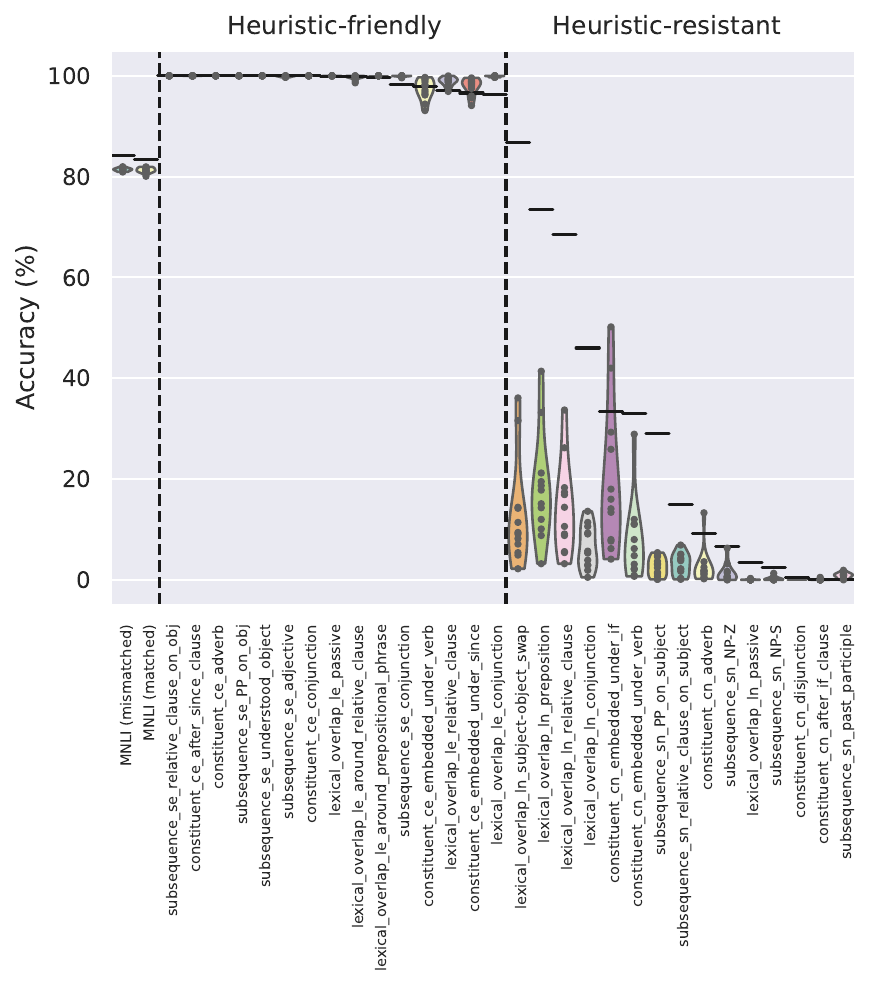}
}
\subfloat[Sparsity $= 70\%$]{
    \includegraphics[width=0.5\linewidth]{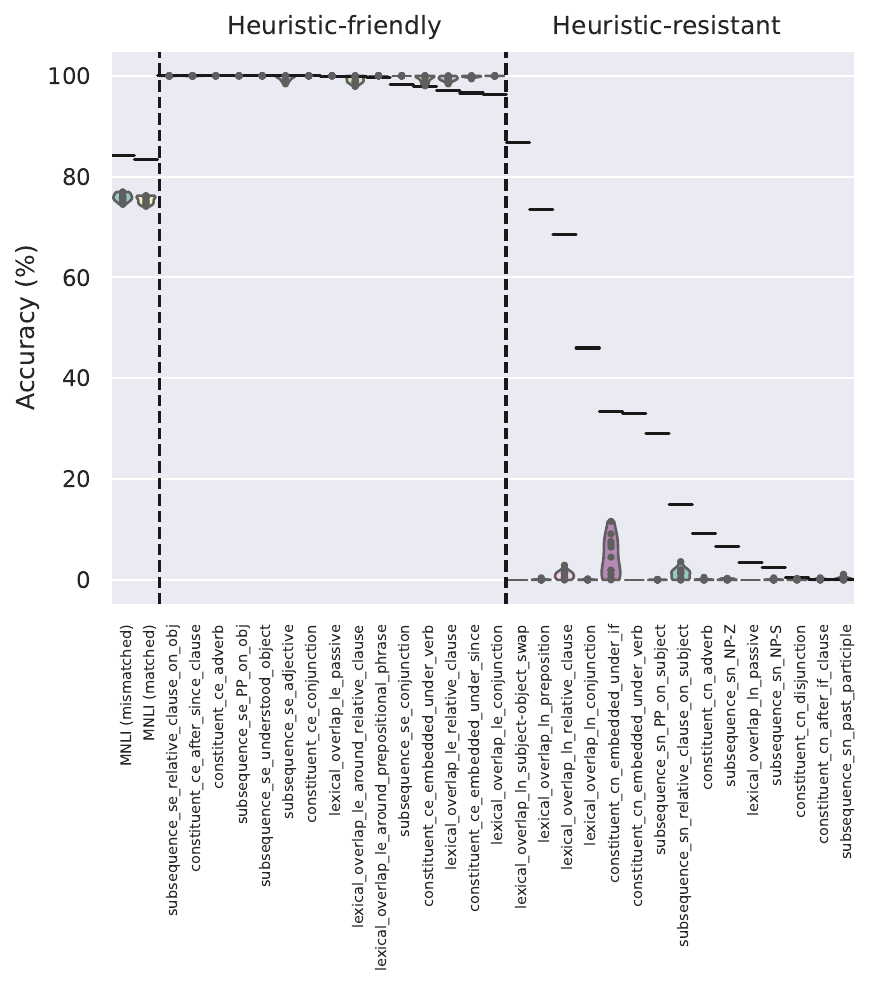}
}
\caption{The full results of the multi-seed pruning experiment on MNLI/HANS. The subnetworks perform close to the model both in-domain and on heuristic-friendly subcases of HANS. Ay $50\%$ sparsity, they show disparate generalization to the heuristic-resistant subcases, while at $70\%$, they degenerate to heuristic behavior. This is also corroborated by the fact that they slightly overperform the full-model on the heuristic-friendly cases.}
\label{fig:multiprunefull}
\end{figure*}

\begin{figure*}[t]
\centering
\subfloat[Sparsity $= 50\%$]{
    \includegraphics[width=0.5\linewidth]{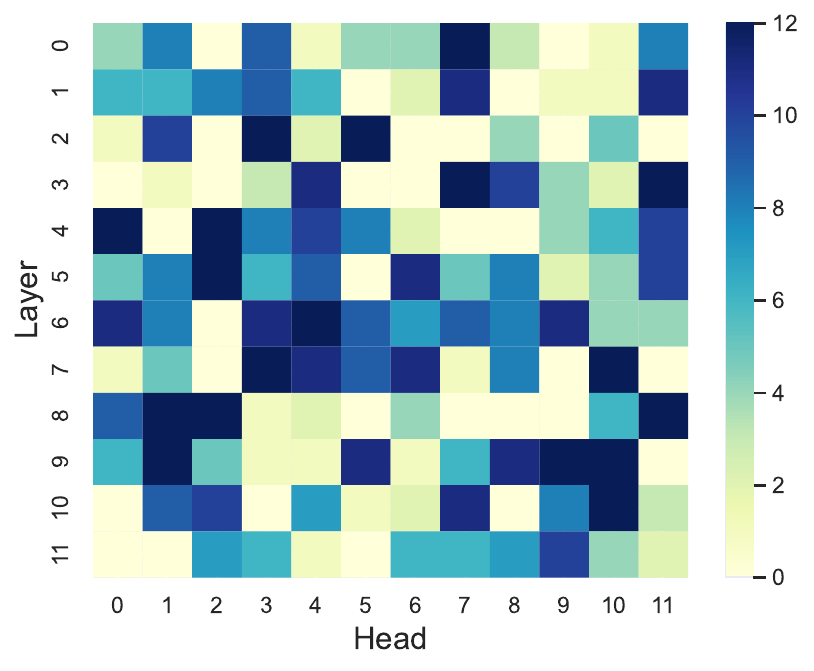}
}
\subfloat[Sparsity $= 70\%$]{
    \includegraphics[width=0.5\linewidth]{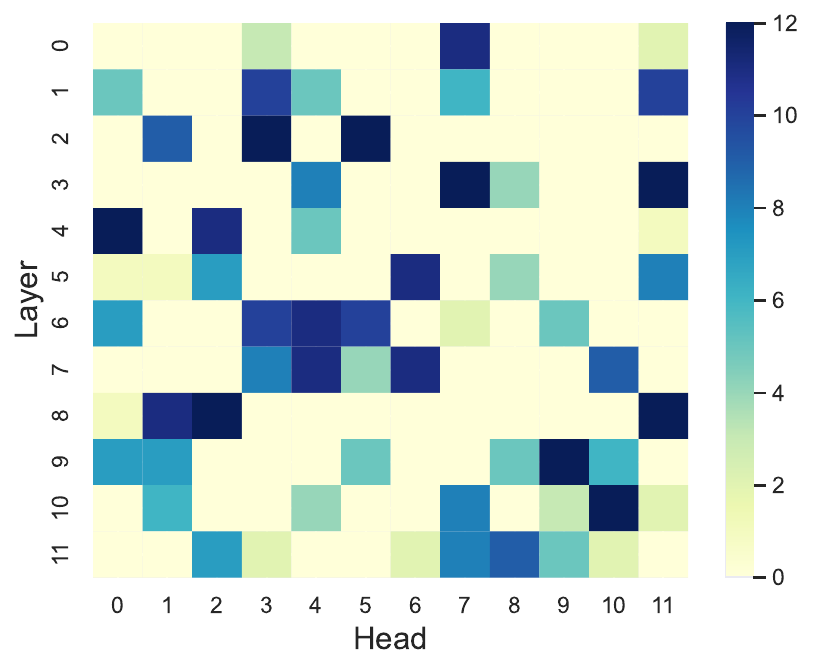}
}
\caption{The frequency of various attention heads across the subnetworks pruned with $12$ random seeds. The heads part of the 70\% sparsity subnetworks (non-generalizing) are also popular in 50\% sparsity (partially generalizing) subnetworks. The Separman's Rho agreement between the frequencies of attention heads at the two sparsities is $0.82$ ($p$-value = $1.6 \cdot 10^{-36}$), corresponding to very strong agreement.
}
\label{fig:membership}
\end{figure*}
\begin{figure*}[t]
\centering
\subfloat[Sparsity $= 30\%$]{
    \includegraphics[width=0.5\linewidth]{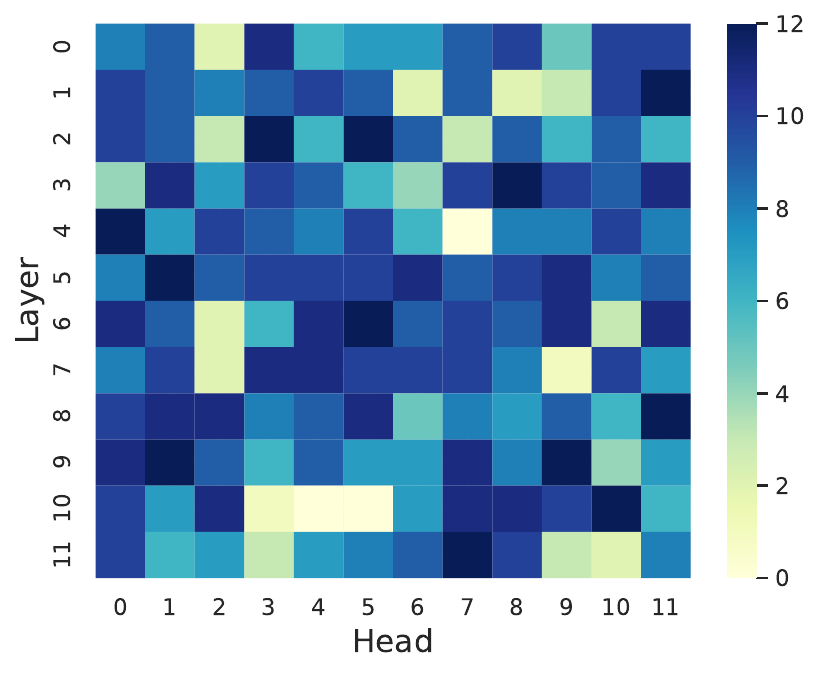}
}
\subfloat[Sparsity $= 60\%$]{
    \includegraphics[width=0.5\linewidth]{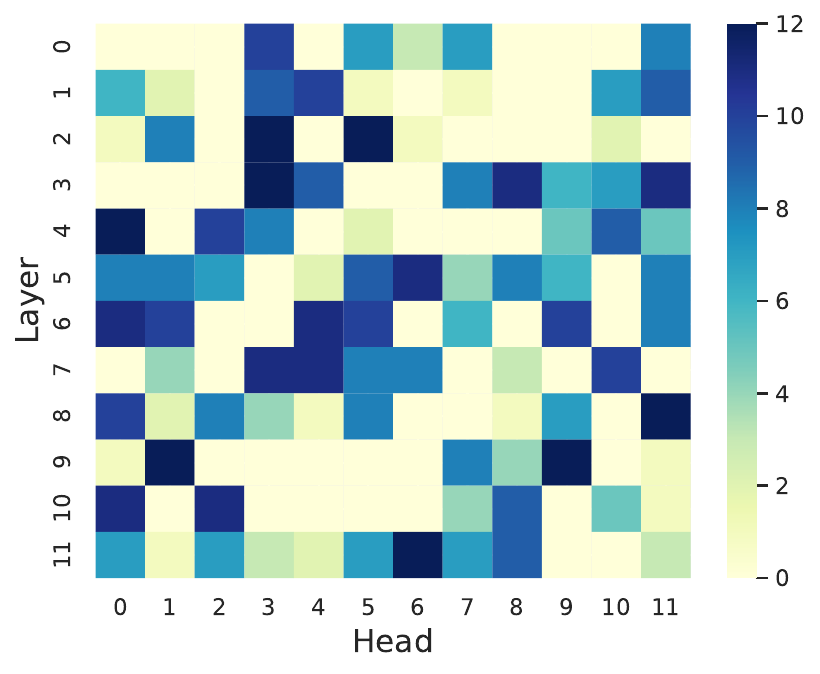}
}
\caption{The frequency of various attention heads across in the QQP subnetworks, pruned with $12$ random seeds. Once again, the frequencies of the heads are highly correlated between the two sparsities. The Separman's Rho agreement in this case is $0.74$ ($p$-value = $7.7 \cdot 10^{-26}$), corresponding to strong agreement.}
\label{fig:membership-qqp}
\end{figure*}

In this section, we provide further exposition on some of the results from the main text.
For these results, we use the full names of the HANS subcases instead of abbreviations.

Figure~\ref{fig:bertgrokfull} showcases the in-domain v/s out-of-domain training dynamics on MNLI/HANS for more subcases.
The trend remains the same---OOD accuracy starts increasing much after ID accuracy saturates.
For the other subcases not shown here, the model never gets off the ground.

Figures~\ref{fig:membership} and~\ref{fig:membership-qqp} show that the frequency of the heads is strongly correlated between the $50\%$ and $70\%$ sparsity subnetworks for MNLI/HANS, and the $30\%$ and $60\%$ sparsity subnetworks for QQP.
This point is further elaborated by the frequency heatmap of Figure~\ref{fig:heatmap-qqp-3060}, the counterpart of Figure~\ref{fig:heatmap-5070}.
Essentially, the heads most important to the heuristic subnetworks also appear often in the generalizing subnetworks, once again showing that the heuristic subnetworks are subsets of generalizing subnetworks.

\section{Results on RoBERTa}
\label{ap:roberta}
\begin{table}
\small
\centering
\begin{tabular}{c c c c c c}
\toprule
\multirow{2}{*}{\textbf{Subcase}} & \multirow{2}{*}{\textbf{Model}} & \multicolumn{4}{c}{\textbf{Subnetworks, 50\% sp.}}\\
\cmidrule{3-6}
& & \textit{Min.} & \textit{Max.} & \textit{Mean} & \textit{STD}\\
\midrule
MNLI (m) & 87.5 & 84.1 & 85.8 & 84.9 & 0.4\\
MNLI (mm) & 87.2 & 84.1 & 85.7 & 84.8 & 0.43\\
\midrule
Prep & 93.2 & 52.2 & 86.4 & 69.1 & 8.9\\
SO-Swap & 99.4 & 46.0 & 87.0 & 67.4 & 12.2\\
Embed-If & 83.1 & 45.6 & 79.8 & 62.2 & 11.0\\
Embed-Verb & 72.6 & 42.3 & 75.6 & 60.4 & 9.6\\
\bottomrule
\end{tabular}
\caption{Different subnetworks at 50\% sparsity show ID performance similar to the full RoBERTa model, but generalize differently.}
\label{tab:roberta_seeds}
\end{table}
\begin{figure*}[t]
\centering
\includegraphics[width=0.2\linewidth]{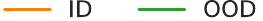}\\
\vspace{-1em}
\subfloat[ID/OOD Accuracy]{
    \includegraphics[width=0.5\linewidth]{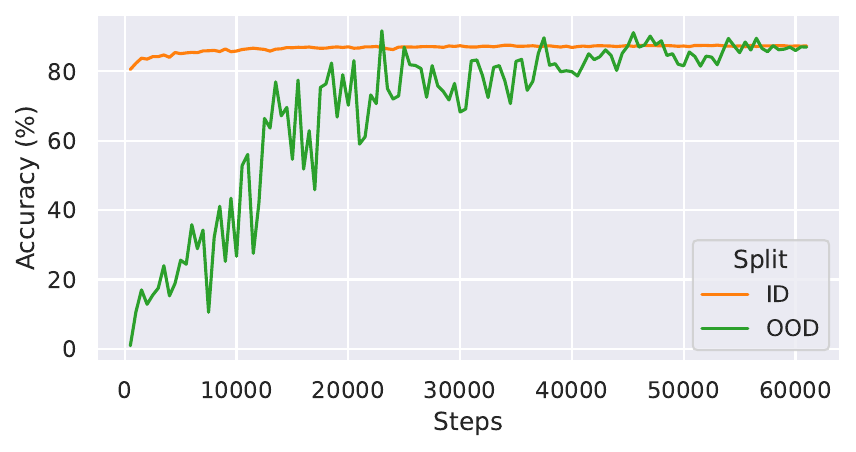}
}
\hfill
\subfloat[Effective Size]{
    \includegraphics[width=0.47\linewidth]{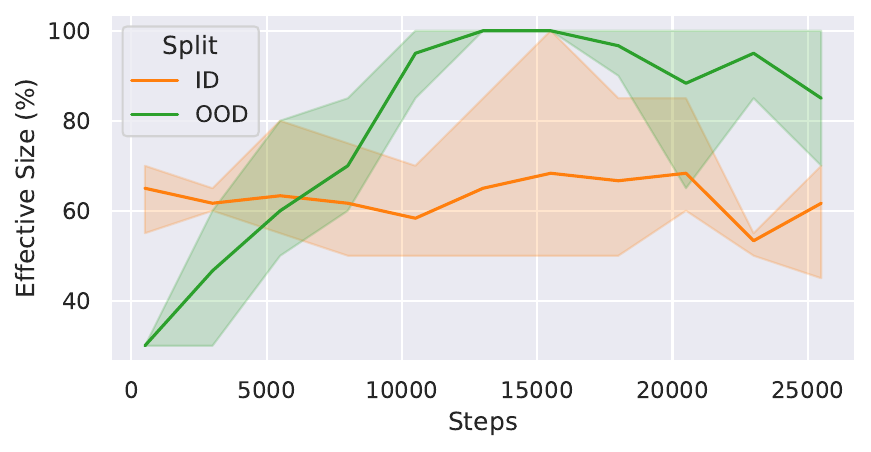}
}
\caption{The accuracy and effective size of the RoBERTa model over training (3 seeds). Generalization is accompanied by an increase in effective size once again. The effective size plot includes three seeds at each sparsity.}
\label{fig:roberta_ae}
\end{figure*}
\begin{table*}
    \centering
    \small
    \begin{tabular}{ll|llll}
        \toprule
        \multirow{2.5}{*}{\textbf{\makecell{Ablated Head}}} & \multicolumn{5}{c}{\textbf{Accuracy (\%)} 
 $\uparrow$}\\
        \cmidrule{2-6}
         & \textbf{\makecell{MNLI (matched)}} & 
\textbf{\makecell{SO-Swap [LO]}} & \textbf{\makecell{Prep [LO]}} & \textbf{\makecell{Embed-If [C]}} & 
\textbf{\makecell{Embed-Verb [C]}}\\
        \midrule
        None & 87.5 & 99.4 & 93.2 & 83.1 & 72.6\\
        \makecell[l]{Avg. (non-HC)} & 87.3$_{\downarrow 0.2}$ & 99.1$_{\downarrow{0.3}}$ & 93.1$_{\downarrow{0.1}}$ & 83.2$_{\uparrow{0.1}}$ & 71.8$_{\downarrow{0.8}}$\\
        \midrule
        Layer 2, Head 4 & 87.5$_{\downarrow 0.0}$ & 95.9$_{\redone{3.5}}$ & 89.8$_{\redone{3.4}}$ & 84.1$_{\blueone{1.0}}$ & 62.5$_{\redtwo{10.1}}$\\
        Layer 3, Head 4 & 87.2$_{\downarrow 0.3}$ & 95.3$_{\redone{4.1}}$ & 86.8$_{\redtwo{6.4}}$ & 80.8$_{\redone{2.3}}$ & 62.6$_{\redtwo{10.0}}$\\
        Layer 4, Head 9 & 86.9$_{\downarrow 0.6}$ & 98.8$_{\downarrow{0.6}}$ & 93.5$_{\uparrow{0.3}}$ & 84.0$_{\uparrow{0.9}}$ & 70.8$_{\redone{1.8}}$\\
        Layer 5, Head 1 & 87.3$_{\downarrow 0.2}$ & 98.5$_{\downarrow{0.9}}$ & 91.8$_{\redone{1.4}}$ & 75.5$_{\redtwo{7.6}}$ & 78.9$_{\bluetwo{6.3}}$\\
        Layer 5, Head 8 & 86.9$_{\downarrow 0.6}$ & 98.9$_{\downarrow{0.5}}$ & 94.6$_{\blueone{1.4}}$ & 76.4$_{\redtwo{6.7}}$ & 60.7$_{\redthree{11.9}}$\\
        Layer 6, Head 2 & 87.5$_{\downarrow 0.0}$ & 98.1$_{\redone{1.3}}$ & 91.5$_{\redone{1.7}}$ & 77.3$_{\redtwo{5.8}}$ & 62.0$_{\redthree{10.5}}$\\
        Layer 6, Head 10 & 87.1$_{\downarrow 0.4}$ & 99.0$_{\downarrow 0.4}$ & 94.2$_{\blueone{1.0}}$ & 80.5$_{\redone{2.6}}$ & 71.7$_{\downarrow 0.9}$\\
        Layer 6, Head 11 & 87.5$_{\downarrow 0.0}$ & 89.8$_{\redthree{9.6}}$ & 73.9$_{\redultra{29.3}}$ & 80.4$_{\redone{2.7}}$ & 59.3$_{\redthree{13.3}}$\\
        Layer 7, Head 9 & 87.3$_{\downarrow 0.2}$ & 99.6$_{\uparrow 0.2}$ & 95.2$_{\blueone{2.0}}$ & 87.2$_{\blueone{4.1}}$ & 62.6$_{\redthree{10.0}}$\\
        Layer 7, Head 10 & 87.1$_{\downarrow 0.4}$ & 98.4$_{\redone{1.0}}$ & 89.9$_{\redtwo{3.3}}$ & 77.6$_{\redtwo{5.5}}$ & 58.9$_{\redthree{13.7}}$\\
        Layer 8, Head 3 & 87.3$_{\downarrow 0.2}$ & 99.3$_{\downarrow 0.1}$ & 93.1$_{\downarrow 0.1}$ & 87.1$_{\blueone{4.0}}$ & 68.5$_{\redone{3.9}}$\\
        \bottomrule
    \end{tabular}
    \caption{The heuristic core of RoBERTa, and the ablation effects of its attention heads. Despite not generalizing on their own, these cause much larger ablation effects than random other heads---head 6.11 shows the maximum effect.}
    \label{tab:roberta_ablations}
\end{table*}

In this section, we verify that our findings extend to a RoBERTa~\citep{liu2019roberta} (\texttt{roberta-base}) model.
We follow the same recipes as the main text, using the same hyperparameters as Appendices~\ref{app:model_details} and~\ref{sec:l0} for fine-tuning and pruning.
Once again, we evaluate on the subcases SO-Swap, Prep, Embed-If, and Embed-Verb in the MNLI-HANS setting.

Table~\ref{tab:roberta_seeds} demonstrates that different subnetworks at $50\%$ sparsity still generalize differently despite performing similarly to the full model in-domain.
Figure~\ref{fig:roberta_ae} tells us that the effective size again goes up along with generalization.
The $12$ subnetworks share $11$ attention heads that also appear in the effective size subnetworks before generalization.
We list these heads along with the effect of ablating them in Table~\ref{tab:roberta_ablations}.
The effects are smaller but consistent with the findings on the BERT model: these heads have much larger ablation effects than random other heads despite not generalizing on their own.
\section{Results on GPT-2}
\label{ap:gpt2}
\begin{table*}
\small
\centering
\begin{tabular}{l c c c c c c c c c}
\toprule
\multirow{2}{*}{\textbf{Subcase}} & \multirow{2}{*}{\textbf{Model}} & \multicolumn{4}{c}{\textbf{Subnetworks, 40\% sp.}} & \multicolumn{4}{c}{\textbf{Subnetworks, 60\% sp.}}\\
\cmidrule(l{2pt}r{2pt}){3-6}\cmidrule(l{2pt}r{2pt}){7-10}
& & \textit{Min.} & \textit{Max.} & \textit{Mean} & \textit{STD} & \textit{Min.} & \textit{Max.} & \textit{Mean} & \textit{STD}\\
\midrule
\textbf{\textit{In-domain}}\\
\quad MNLI (matched) & 82.0 & 78.7 & 80.3 & 79.4 & 0.5 & 74.3 & 76.9 & 75.5 & 0.8\\
\quad MNLI (mismatched) & 82.2 & 79.3 & 80.8 & 80.1 & 0.5 & 74.4 & 77.5 & 76.3 & 1.1\\
\midrule
\textbf{\textit{Out-of-domain (contradiction)}}\\
\quad Embed-If [C] & 47.4 & 17.8 & 36.3 & 26.9 & 5.8 & 16.8 & 51.9 & 33.9 & 12.2\\
\midrule
\textbf{\textit{Out-of-domain (entailment)}}\\
\quad Rel-Clause [S] & 100.0 & 96.4 & 99.7 & 98.1 & 1.1 & 66.3 & 97.8 & 85.3 & 9.8\\
\quad Embed-Since [C] & 88.8 & 85.0 & 98.6 & 92.0 & 4.2 & 57.9 & 92.0 & 79.9 & 11.7\\
\bottomrule
\end{tabular}
\caption{The performance of 12 different pruned subnetworks of GPT-2 at $40\%$ and $60\%$ sparsity. We still see massive variance out-of-domain compared to in-domain evaluation but the performance on the `contradiction' examples (OOD) goes up with sparsity, and that of `entailment' (OOD-E) goes down.}
\label{tab:gpt2_seeds}
\end{table*}
\begin{figure*}[t]
\centering
\includegraphics[width=0.3\linewidth]{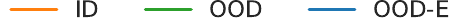}\\
\vspace{-1em}
\subfloat[ID/OOD Accuracy]{
    \includegraphics[width=0.5\linewidth]{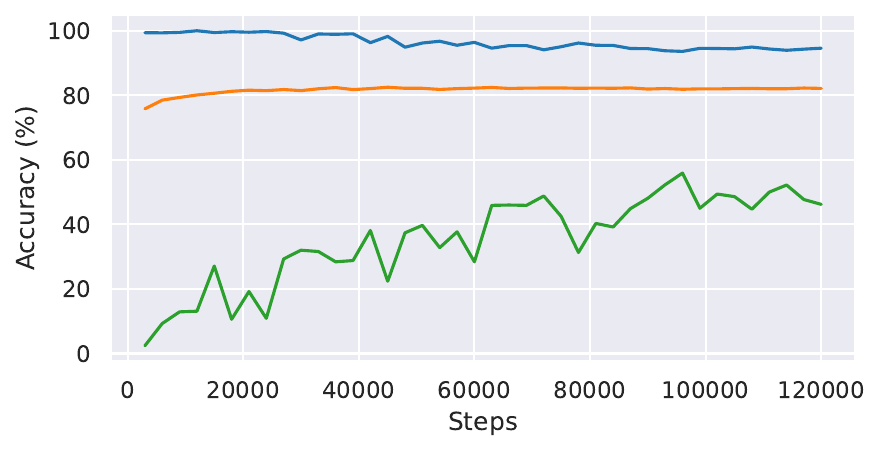}
}
\hfill
\subfloat[Effective Size]{
    \includegraphics[width=0.47\linewidth]{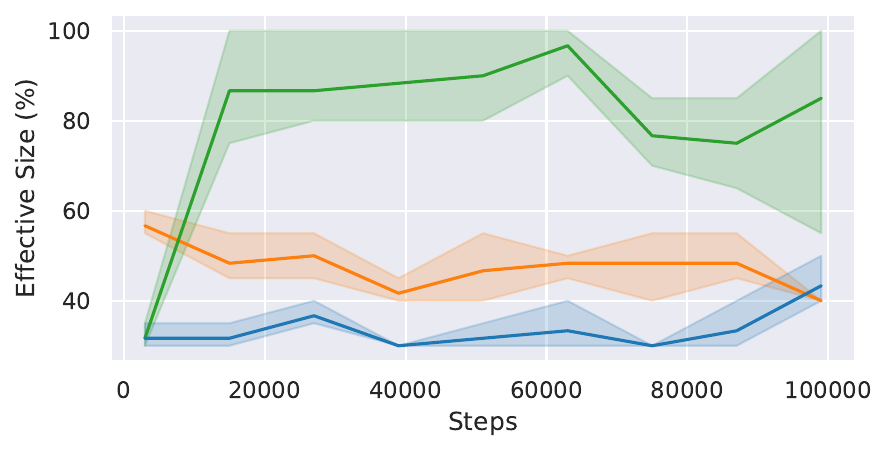}
}
\caption{The accuracy and effective size of the GPT-2 model over training (3 seeds). The effective size on OOD increases with generalization. Interestingly, the accuracy on the ``entailment'' OOD (OOD-E) cases decays slowly later into the training, and is accompanied by a slight increase in the corresponding effective size.}
\label{fig:gpt2_ae}
\end{figure*}
\begin{table*}
    \centering
    \small
    \begin{tabular}{lllll}
        \toprule
        \multirow{4.5}{*}{\textbf{\makecell{Ablated Head}}} & \multicolumn{4}{c}{\textbf{Accuracy (\%)} 
 $\uparrow$}\\
        \cmidrule{2-5}
         & \multicolumn{1}{c}{\textit{ID}} & \multicolumn{1}{c}{\textit{OOD}} & \multicolumn{2}{c}{\textit{OOD-E}}\\
         \cmidrule(l{2pt}r{2pt}){2-2}
         \cmidrule(l{2pt}r{2pt}){3-3}
         \cmidrule(l{2pt}r{2pt}){4-5}
         & \makecell{MNLI (matched)} & 
\makecell{Embed-If [C]} & \makecell{Rel-Clause [S]} & \makecell{Embed-Since [C]}\\
        \midrule
        None & 82.0 & 47.4 & 100.0 & 88.8\\
        \makecell[l]{Avg. (non-HC)} & 81.9$_{\downarrow 0.1}$ & 47.2$_{\downarrow{0.2}}$ & 100.0$_{\downarrow{0.0}}$ & 89.0$_{\uparrow{0.2}}$\\
        \midrule
        Layer 0, Head 7 & 81.6$_{\downarrow 0.4}$ & 49.4$_{\blueone{2.0}}$ & 100.0$_{\downarrow{0.0}}$ & 88.5$_{\downarrow{0.3}}$\\
        Layer 0, Head 9 & 81.8$_{\downarrow 0.2}$ & 46.8$_{\downarrow{0.6}}$ & 100.0$_{\downarrow{0.0}}$ & 88.2$_{\downarrow{0.6}}$\\
        Layer 1, Head 0 & 81.2$_{\downarrow 0.8}$ & 34.5$_{\redthree{12.9}}$ & 100.0$_{\downarrow{0.0}}$ & 92.4$_{\blueone{3.6}}$\\
        Layer 1, Head 1 & 81.9$_{\downarrow 0.1}$ & 48.2$_{\uparrow{0.8}}$ & 100.0$_{\downarrow{0.0}}$ & 89.1$_{\uparrow{0.3}}$\\
        Layer 3, Head 0 & 81.9$_{\downarrow 0.1}$ & 51.6$_{\blueone{4.2}}$ & 99.9$_{\downarrow{0.1}}$ & 87.1$_{\redone{1.7}}$\\
        Layer 4, Head 9 & 81.4$_{\downarrow 0.6}$ & 34.5$_{\redthree{12.9}}$ & 100.0$_{\downarrow{0.0}}$ & 91.9$_{\blueone{3.1}}$\\
        Layer 4, Head 11 & 80.9$_{\redone{1.1}}$ & 54.5$_{\bluetwo{7.1}}$ & 100.0$_{\downarrow{0.0}}$ & 88.4$_{\downarrow{0.4}}$\\
        Layer 5, Head 5 & 80.7$_{\redone{1.3}}$ & 57.6$_{\bluethree{10.2}}$ & 99.5$_{\downarrow{0.5}}$ & 85.1$_{\redone{3.7}}$\\
        Layer 5, Head 8 & 81.5$_{\downarrow 0.5}$ & 44.2$_{\redone{3.2}}$ & 100.0$_{\downarrow{0.0}}$ & 89.4$_{\uparrow{0.6}}$\\
        Layer 6, Head 3 & 81.2$_{\downarrow 0.8}$ & 40.6$_{\redtwo{6.8}}$ & 100.0$_{\downarrow{0.0}}$ & 90.4$_{\blueone{1.6}}$\\
        Layer 6, Head 7 & 81.0$_{\redone{1.0}}$ & 52.5$_{\bluetwo{5.1}}$ & 100.0$_{\downarrow{0.0}}$ & 88.5$_{\downarrow{0.3}}$\\
        Layer 6, Head 11 & 81.5$_{\downarrow 0.5}$ & 44.5$_{\redone{2.9}}$ & 100.0$_{\downarrow{0.0}}$ & 89.9$_{\blueone{1.1}}$\\
        \bottomrule
    \end{tabular}
    \caption{The heads common to all $40\%$ and $60\%$ sparsity subnetworks of GPT-2, and their ablation effects. On Embed-If, we see that some heads (4.11, 5.5, 6.7) show an increased accuracy on ablation and some (1.0, 4.9, 6.3) a strong decrease. On the `entailment' subcases, the ablation effects of individual heads is generally small.}
    \label{tab:gpt2_ablations}
\end{table*}

In this subsection, we ask if our findings and explanations can extend to decoder-only models.
We work with a GPT-2 small (117M) model~\citep{radford2019language}.
Most of the hyperparameters for both fine-tuning and pruning are the same as those listed in Appendices~\ref{app:model_details} and~\ref{sec:l0}.
However, we found that the GPT-2 model, even as a whole, generalizes very poorly to the HANS subcases.
After training for $10$ epochs ($12270$ steps), \texttt{constituent\_cn\_embed\_under\_if} is the only ``contradiction'' subcase with an accuracy above $30\%$. 
Hence, we select this setting, and Embed-If as the only adversarial out-of-domain subcase.
Interestingly, however, we observed that the accuracies of some of the \emph{heuristic-friendly} subcases started dropping later into the training.
We select \texttt{subsequence\_se\_relative\_clause\_on\_obj} (Rel-Clause [SE]) and \texttt{constituent\_ce\_embedded\_under\_since} (Embed-Since [CE]) as two representative subcases and group them under the split OOD-E (OOD-Entailment).
We perform the experiments for both the OOD and OOD-E subcases.

We evaluate 12 pruned subnetworks at $40\%$ and $60\%$ sparsity in Table~\ref{tab:gpt2_seeds}.
Like before, the subnetworks perform similarly in-domain but generalize differently.
However, one striking feature stands out.
As we increase the sparsity, the average performance on the `contradiction' OOD cases \emph{increases}, whereas that of the `entailment' cases \emph{decreases}.

One way to reconcile these results is for some heuristics to be \emph{contradiction} heuristics, and for the model to develop further non-heuristic heads around a \emph{contradiction} heuristic core.
Indeed, in Figure~\ref{fig:gpt2_ae}, we note two interesting trends: (1) the model accuracy decreases slightly on the OOD-E subcases late into training and (2) a mild increase in the OOD-E effective size accompanies this decrease in accuracy.
A possible explanation for the latter could be that some heuristics for spotting `contradiction' are also learned, and play an analogous role in the heuristic core.

We find the heads common to all $40\%$ and $60\%$ subnetworks and tabulate their ablation effects in Table~\ref{tab:gpt2_ablations}.
We note the following: (1) On Embed-If, some heads show a strong increase in accuracy on ablation, and some a strong decrease. 
This would be consistent with some heads implementing entailment heuristics and some contradiction heuristics.
On the other hand, (2) None of the heads show large ablation effects on Rel-Clause. On Embed-Since---the other entailment subcase---some heads show small but 
significant effects.

Therefore, our results on the OOD subcases are consistent with the model implementing both entailment and contradiction heuristics.
On the other hand, some questions remain regarding the OOD-E subcases in GPT-2---namely, why the performance of the model tapers off late into the training, and why there is a variance in performance at \emph{high} sparsities (in contrast to the results on BERT) despite none of the common heads having significant ablation effects.
We leave it to future work to answer these questions.

\end{document}